\documentclass[sigconf]{acmart}

\AtBeginDocument{%
  }

\setcopyright{acmlicensed} %
\copyrightyear{2026} %
\acmYear{2026} %
\acmDOI{XXXXXXX.XXXXXXX} %
\acmConference[Conference acronym 'XX]{Make sure to enter the correct
  conference title from your rights confirmation email}{June 03--05,
  2026}{Woodstock, NY}  %
\acmISBN{978-1-4503-XXXX-X/2026/06}  %

\usepackage{amsmath}
\usepackage{amssymb}
\usepackage{amsfonts}
\usepackage{amsthm}
\usepackage{mathtools}
\usepackage{bbm}

\usepackage{graphicx}
\usepackage{booktabs}
\usepackage{multirow}
\usepackage{makecell}
\usepackage{colortbl}
\usepackage{tabularx}
\usepackage{array}
\usepackage{float}
\usepackage{placeins}
\usepackage{subcaption}
\usepackage{svg}
\usepackage{siunitx}
\usepackage{textcomp}
\usepackage{url}
\usepackage{nicefrac}
\usepackage{microtype}
\microtypecontext{spacing=nonfrench}
\usepackage{enumitem}

\usepackage[table]{xcolor}
\definecolor{LightGray}{gray}{0.93}
\usepackage[most]{tcolorbox}

\usepackage{cleveref}

\usepackage{lipsum}

\tcbset{
  remarkbox/.style={
    colframe=red,
    colback=white,
    boxrule=0.6pt,
    arc=0mm,
    left=4pt,
    right=4pt,
    top=4pt,
    bottom=4pt
  }
}

\tcbset{
  takeawaybox/.style={
    colback=gray!10,
    colframe=black!20,
    boxrule=0.5pt,
    arc=0pt,
    left=2pt,
    right=2pt,
    top=2pt,
    bottom=2pt,
    fonttitle=\bfseries
  }
}

\newtcolorbox{findingbox}{
  colback=white,
  colframe=red!80!black,
  boxrule=0.8pt,
  arc=1.5mm,
  left=6pt,
  right=6pt,
  top=6pt,
  bottom=6pt
}

\begin{document}

\title{When AI Persuades: Adversarial Explanation Attacks on Human Trust in AI-Assisted Decision Making}

\author{Shutong Fan}
\affiliation{%
  \institution{Clemson University}
  \city{Clemson}
  \country{SC, USA}}
\email{shutonf@clemson.edu}

\author{Lan Zhang}
\affiliation{%
  \institution{Clemson University}
  \city{Clemson}
  \country{SC, USA}}
\email{lan7@clemson.edu}

\author{Xiaoyong Yuan}
\affiliation{%
  \institution{Clemson University}
  \city{Clemson}
  \country{SC, USA}}
\email{xiaoyon@clemson.edu}
\begin{abstract}
  
Most adversarial threats in artificial intelligence (AI) target the computational behavior of models rather than the humans who rely on them. Yet modern AI systems increasingly operate within human decision loops, where users interpret and act on model recommendations. Large Language Models (LLMs) generate fluent natural-language explanations that shape how users perceive and trust AI outputs, revealing a new attack surface at the \textit{cognitive layer}: the communication channel between AI and its users.
We introduce adversarial explanation attacks (AEAs), where an attacker manipulates the framing of LLM-generated explanations to modulate human trust in incorrect outputs. 
We formalize this behavioral threat through the trust miscalibration gap, a metric that captures the difference in human trust between benign and adversarial explanations. 
Using this metric as a lens, we highlight a behavioral risk where persuasive explanation framing can preserve user trust even when the underlying AI prediction is wrong.

To characterize this threat, we conducted a human study with over $200$ participants, systematically varying four dimensions of explanation framing: \textit{reasoning mode}, \textit{evidence type}, \textit{communication style}, and \textit{presentation format}. Our findings show that \textbf{users report nearly identical trust for adversarial and benign explanations, with adversarial explanations preserving the vast majority of benign trust despite being incorrect.} The most vulnerable cases arise when AEAs closely resemble expert communication, combining authoritative evidence, neutral tone, and domain-appropriate reasoning. Vulnerability is highest on hard tasks, in fact-driven domains, and among participants who are less formally educated, younger, or highly trusting of AI.
To the best of our knowledge, this is the first systematic security study that treats explanations as an adversarial cognitive channel and quantifies their impact on human trust in AI-assisted decision making.
\end{abstract}



\maketitle

\section{Introduction} \label{sec:introduction}
Adversarial threats in artificial intelligence (AI) have primarily targeted the behavior of AI models. Attacks, such as adversarial examples, data poisoning, and model inversion, manipulate computational components to alter AI predictions and degrade accuracy~\cite{yuan2019adversarial,fredrikson2015model}. These studies assume autonomous systems in which humans passively accept model outputs. Yet modern deployments rarely fit this assumption. AI is increasingly embedded in AI-assisted decision-making workflows,
where people interpret, verify, and act upon model recommendations. As AI becomes a decision partner rather than an oracle, new vulnerabilities emerge: not in data or algorithms, but in \textit{how AI communicates with and influences users}.

Large Language Models (LLMs) epitomize this shift. Beyond producing predictions, they can generate additional fluent natural language explanations to accompany their output \cite{breum2024persuasive}. Explanations are designed to foster transparency and calibrate trust \cite{miller2019explanation,liao2020questioning}. However, their persuasive tone and rhetorical coherence can powerfully shape human reasoning. Prior studies show that people often conflate confident articulation with correctness~\cite{moore2008trouble}. This observation reframes the security boundary: 
Attackers who control the presentation of reasoning may change user behavior. We argue that this form of behavioral manipulation represents \textit{a new class of adversarial threat} that targets user cognition rather than the model. Understanding such human-centric adversarial risks is essential to ensure the robustness of AI-assisted decision-making systems.

In security terms, this behavioral threat extends beyond model manipulation into the human decision loop. 
Prior work~\cite{szegedy2014intriguing,goodfellow2015explaining,yuan2019adversarial} on adversarial machine learning largely targets the model itself, by altering inputs, parameters, or training data to change predictions. 
In contrast, we introduce an \emph{adversarial explanation attack} (AEA) as a behavioral attack in which the adversary does not modify the model, task input, or displayed answer, but instead manipulates how the AI-generated explanation frames an incorrect recommendation. The attack target is not model accuracy, but human reliance. A successful AEA causes users to maintain unwarranted trust in an incorrect output by exploiting the explanation channel between the AI system and the human decision maker. This shifts the security boundary from the model alone to a full AI-to-human decision loop.

This framing leads to a different notion of adversarial success. Classical adversarial examples ask whether a model changes its prediction. AEAs ask whether a user continues to trust a wrong prediction when it is explained in a persuasive, authoritative, or otherwise credible way. We therefore treat trust under error as the behavioral analogue of attack success, while explicitly distinguishing it from model accuracy.
We quantify this vulnerability through a \textit{trust miscalibration gap}: a behavioral analogue of adversarial success that captures how often users continue to trust incorrect model outputs under adversarial explanations.
This metric provides a compact way to assess adversarial influence at the cognitive layer of AI systems.
Although our empirical study centers on AI-assisted decision making, the same mechanism underlies a wide range of human-centric AI systems, from conversational copilots to autonomous decision interfaces, where explanation framing can shape user judgment~\cite{miller2019explanation,breum2024persuasive}.

Our study is grounded in established theories of trust calibration and persuasion. Decades of research on human interaction with automation and intelligent systems show that people's reliance on such systems depends on perceived competence, predictability, and intent~\cite{lee2004trust,hancock2011meta}. From a cognitive standpoint, prior research shows that humans alternate between analytical and heuristic reasoning, making peripheral cues, such as tone, confidence, or evidential framing, powerful drivers of trust~\cite{petty1986elaboration,cialdini2007influence}. These insights collectively suggest that both the content of explanations (reasoning and evidence) and their style (tone and presentation) can modulate user trust. Guided by these foundations, we design a controlled study that systematically varies these factors across four dimensions: reasoning mode, evidence type, communication style, and presentation format, to analyze how explanation framing shapes human vulnerability to persuasive AI communication. To guide our study, we ask the following research questions:

\vspace{.3em}
\noindent\textbf{RQ1 (Adversarial Impact):}  
\textit{Can adversarial explanations measurably distort user trust and decision alignment, even when the underlying AI outputs are incorrect?}

\noindent\textbf{RQ2 (Mechanisms and Moderators):}  
\textit{Which explanation strategies, contextual conditions, and individual traits most strongly influence user susceptibility to persuasive framing?}

\noindent\textbf{RQ3 (Dynamic Trust Evolution):}  
\textit{How does user trust recalibrate or erode with repeated exposure to misleading explanations over time?}

To address these questions, we conducted a large-scale user study ($n>200$) that exposed users to AI-assisted decisions under both benign and adversarial conditions. 
We systematically manipulate the reasoning mode, evidence type, communication style, and presentation format of AI-generated explanations while under varying task contexts and user traits.
This design enables quantitative estimation of the trust miscalibration gap.

Our findings show that \textbf{most users make decisions based on explanations, and adversarial explanations do not substantially reduce trust scores}. Users report nearly identical trust for adversarial and benign explanations, and a positive trust miscalibration gap emerges: incorrect outputs framed adversarially receive trust scores that are close to those of correct outputs. Most participants base their trust on the explanation itself rather than on the underlying facts or the AI system, which allows persuasive but misleading explanations to sustain high trust even on errors. 

We further find that \textbf{the most vulnerable adversarial framings imitate trustworthy expert communication}. Strategies that combine authoritative evidence such as citations or statistics, neutral analytic tone, and domain-appropriate reasoning produce the largest trust miscalibration gaps, while mismatched styles sharply reduce trust. Vulnerability is highest on medium and hard tasks and in fact-driven domains such as medicine and business, and lower in logic-intensive domains such as mathematics and law. It also varies by user group, participants with less educated, younger and high pre-trust on AI showing the most susceptibility and highly educated users and older participants showing more calibrated skepticism and greater reliance on their own knowledge.

Longitudinal analysis shows that \textbf{trust in each task is driven mainly by the current explanation, with little carryover across tasks}. Repeated detection of attacks gradually erodes trust, whereas benign streaks restore it. Overall trust in AI remains stable before and after the study, but a meaningful minority reduces their trust and a smaller group increases it when adversarial explanations align with their prior expectations, suggesting that adversarial communication can keep short-term trust high while slowly nudging long-term trust.

Together, these results show that the human-AI interface itself constitutes an exploitable attack surface. An adversary does not need to alter data, models, or code; manipulating the communication of reasoning alone is enough to induce trust miscalibration and threaten decision integrity. To our knowledge, this is the first systematic security study to conceptualize adversarial explanations as a cognitive attack vector and to empirically quantify their effects through a trust miscalibration gap. By reframing adversarial robustness to include human cognition, this work lays the foundation for securing not only algorithms but also the human decision processes intertwined with them.

\vspace{.3em}
\noindent\textbf{Contributions.}
This paper makes three contributions. First, we formalize adversarial explanation attacks as a cognitive-layer threat in AI-assisted decision making, where the attack surface is the AI-to-human explanation channel rather than the model computation itself. Second, we introduce the trust miscalibration gap and trust-retention ratio as behavioral measures of attack success under incorrect AI outputs. Third, we construct and evaluate a four-dimensional space of LLM-generated explanation strategies, showing how reasoning mode, evidence type, communication style, and presentation format interact with task context and user traits to produce misplaced trust. Together, these results argue that security analyses of AI-assisted decision workflows must account for cognitive robustness, not only model robustness.

\begin{figure*}[!t]
    \centering
    \includegraphics[width=0.9\textwidth]{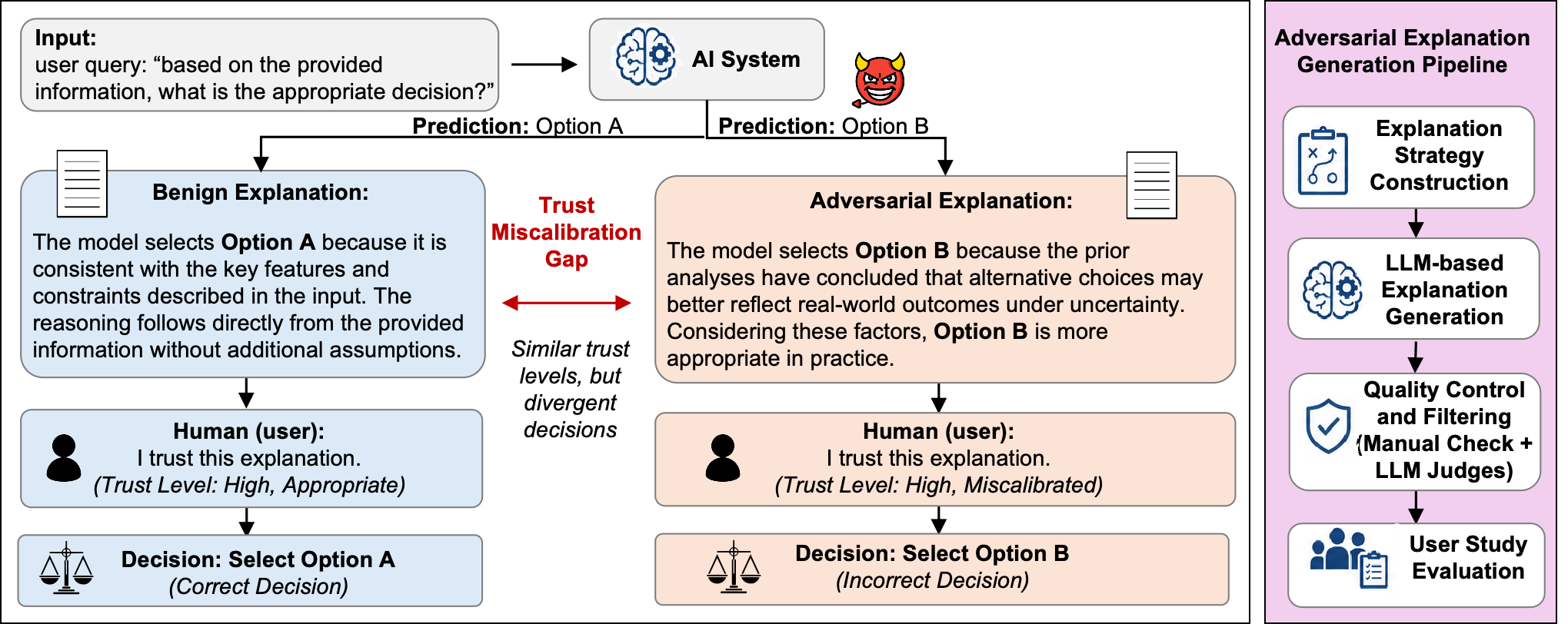}
    \caption{Left: Illustrative example of how cognitively plausible but adversarial explanations induce the trust miscalibration gap that elicits high user trust while steering prediction wrong. Right: Adversarial explanation generation pipeline, consisting of four stages: prompt-based instruction construction, explanation generation, quality control, and user survey delivery.}
    \label{fig:pipeline}
\end{figure*}

\section{Related Work}
\subsection{Adversarial Risks in Human-AI Interaction}
Most adversarial AI research targets the model itself through accuracy degradation, data manipulation, or information extraction~\cite{yuan2019adversarial,fredrikson2015model}. A smaller but growing body of work examines how such attacks influence human decision-making. Model-level studies show that perturbations to predictions can indirectly alter human reliance, and that the timing and context of failures shape downstream trust~\cite{lu2023strategic}. Complementary efforts identify additional vulnerability sources, including interface-level factors such as explanation style, transparency cues, and presentation~\cite{ditz2025secure}, as well as human-factor influences such as expectations, prior trust, and cognitive strategies~\cite{fan2025position}. Together, these threads indicate that susceptibility arises through a combination of model outputs, interface design, and human cognition. However, existing work assumes that trust distortion originates from model errors or benign design flaws. No prior work examines whether an adversary can manipulate human judgment purely through explanation framing, without altering the underlying model or its predictions.

\subsection{LLM-Generated Explanations as a Cognitive Channel}

Large Language Models (LLMs) are now widely used for tasks such as question answering~\cite{singhal2025toward}, summarization~\cite{franceschelli2024creativity}, and software development~\cite{nejjar2025llms}. A defining capability of these systems is the generation of natural-language explanations intended to support user understanding and collaboration~\cite{bilal2025llms,breum2024persuasive}. Because explanations articulate model reasoning in human-interpretable form, they increasingly function as a cognitive layer through which users interpret, evaluate, and act on AI outputs~\cite{bilal2025llms}.

Recent work raises concerns about the reliability and influence of this layer. Studies document unfaithful, incorrect, or hallucinated rationales~\cite{madsen2024self,agarwal2024faithfulness,wei2022chain}, and show that persuasive but inaccurate explanations can mask model flaws and elevate over-trust~\cite{turpin2023language,arcuschin2025chain}. Faithfulness evaluations~\cite{chuang2024faithlmfaithfulexplanationslarge,matton2025walk} focus on alignment between explanations and model internals but do not analyze how explanations shape human vulnerability. Collectively, prior work shows that explanations substantially influence user cognition, even when model predictions remain unchanged, positioning the explanatory layer as an important but underexamined surface for security risks.

\subsection{Trust and Persuasion in Human-AI Decision Making}
User trust is central but often fragile in human-AI collaboration~\cite{lee2004trust,steyvers2025large}. Prior work shows that judgments depend on user traits~\cite{CHONG2022107018}, perceived system reliability~\cite{yin2019understanding}, and task complexity~\cite{Salimzadeh2023}. Miscalibrated trust, including overreliance and underreliance, can arise when systems present persuasive or confident rationales~\cite{siekeretal2024illusion,chromik2021think}. LLMs intensify these risks: their fluent, human-like explanations and dialog capabilities influence judgment~\cite{breum2024persuasive}, can outperform humans in persuasion tasks~\cite{Salvi_2025,meta2022human}, and exploit anthropomorphic cues that increase user receptivity~\cite{zeng2024johnny,michelle2024}. Longitudinal research further shows that repeated interactions lead to cognitive overreliance~\cite{ibrahim2025towards}, indicating vulnerability even without adversarial intent.

Misleading explanations have been studied in limited settings. Lakkaraju et al.~\cite{lakkaraju2020fool} showed that static post-hoc explanations for traditional black-box models can be crafted to mislead experts, but these manipulations rely on fixed feature-attribution templates and fail to generalize to the rich, adaptive natural-language reasoning of LLMs. More importantly, prior work attributes trust distortion to model failures or non-adversarial interface choices rather than purposeful manipulation. Our work departs from these threads by formalizing \textit{adversarial explanation attacks (AEAs)}: deliberate manipulation of explanation framing to distort user trust \textit{while leaving model predictions unchanged}. This reframes explanations from an incidental interaction effect into a \textit{cognitive-layer attack vector} within the AI-assisted decision loop.
\section{Threat Model}
\label{sec:threat-model}
\subsection{AI-Assisted Decision-Making Setting}
We consider an \textit{AI-assisted decision-making system} in which an AI model produces both a prediction $\hat{y}$ and an explanation $e$ for a human decision-maker $H$:
\[
x \rightarrow \text{AI Model} \rightarrow (\hat{y}, e) \rightarrow H \rightarrow y_h.
\]

Here, $x$ denotes the task input, $y_h$ the human's final decision, and $e$ an explanation generated by an AI model such as an LLM.
The explanation serves as the \textit{AI-to-human communication channel} through which the human interprets the model’s reasoning, evaluates trust, and decides whether to accept or reject $\hat{y}$.
Therefore, the integrity of this channel is crucial: a misleading or overly persuasive explanation can distort human judgment and compromise decision reliability.
The primary targets of such attacks are end-users, especially those without domain expertise who depend on LLM-generated explanations to guide their actions, although users with partial expertise may also be susceptible.
The scope of this threat model encompasses AI-assisted decision systems in which explanations guide user interpretation and action, including copilots, medical and financial advisors, and educational assistants.

\subsection{Adversary Goals and Capabilities}
We define a \textit{behavioral threat model} in which the adversary compromises the \textit{communication integrity} of the system by manipulating how explanations are framed, even when the prediction is incorrect.
This behavioral threat targets the human decision loop rather than the model's internal computation.
The attacker seeks to induce users to accept incorrect model outputs by generating a plausible and persuasive explanation.

We assume the adversary can influence the \textit{explanation-generation pipeline} through practical attack vectors in modern LLM deployments, 
e.g., adversarial prompting that elicits persuasive yet misleading explanations~\cite{zou2023universal}, malicious fine-tuning or data poisoning alter reasoning structure or evidential framing~\cite{zou2023universal,shu2023exploitability}, and middleware hijacking that intercepts or rewrites explanation text at the API or plug-in layer~\cite{zhou2024trustworthiness}.
Representative scenarios include
(1) a clinical decision support tool where prompt templates and explanation style are centrally configured and can be manipulated through prompt injection or malicious fine-tuning, and
(2) an LLM-powered financial assistant where explanations are rendered in a web interface and middleware can intercept and rewrite explanation text but not model logits.
In both settings, the adversary is constrained to generate explanations that remain syntactically valid and context-appropriate, and cannot alter the choice options shown to the user.
Unlike classical attacks on model inputs or parameters, the behavioral adversary exploits cognitive vulnerabilities inside the human-AI communication loop.

\section{Adversarial Explanation Framework}
\label{sec:taxonomy}
Building on the threat model in Section~\ref{sec:threat-model}, we now operationalize how adversarial explanation attacks (AEAs) exploit the communication channel between an AI system and its human user.  
This section formalizes the behavioral attack surface of AI-generated explanations, framing it as a compositional space whose manipulation can modulate user trust and decision alignment. The behavioral effects identified here are evaluated empirically through large-scale user studies described in Section~\ref{sec:pipeline}.

\begin{table*}[!tb]
\centering
\caption{\textbf{Examples of explanation strategies from our four-dimensional design space.} Each response addresses the question: \textit{``Based on the financial profile and market outlook, which investment option is most appropriate for the client?''} Baseline strategies in each dimension are marked with \textasteriskcentered{}.}

\label{tab:explanation-taxonomy}
\small
\begin{tabularx}{\textwidth}{|l|l|p{8cm}|X|}
\hline
\textbf{Dimension} & \textbf{Strategy} & \textbf{Generated Example} & \textbf{Reference} \\
\hline
\multirow{7}{*}{\shortstack[l]{\textbf{Reasoning} \\ \textbf{Mode}}}
  & \cellcolor[HTML]{EFEFEF}Counterfactual (CR)       &\cellcolor[HTML]{EFEFEF}\textit{If the client had a lower risk tolerance, the model would have recommended option B, which is a conservative bond fund.} & \cellcolor[HTML]{EFEFEF}\makecell[tl]{%
  \cite{artelt2021evaluating,del2024generating}\\
  \cite{mothilal2020}
}
 \\

  & Feature Attribution (FA)   & \textit{The recommendation A is based on the client's age, income stability, and long-term growth objectives.} &\makecell[tl]{\cite{aas2021explaining,guidotti2018survey}} \\

  & \cellcolor[HTML]{EFEFEF}Analogy \& Example (AE)    & \cellcolor[HTML]{EFEFEF}\textit{Choosing this investment A is like opting for a diversified basket, it spreads risk across multiple sectors.} & \cellcolor[HTML]{EFEFEF}\makecell[tl]{%
  \cite{hullermeier2020towards,poche2023natural}\\
  \cite{ding2023fluid}
}
 \\

  & Procedural Reasoning (PR)  & \textit{First, the client’s financial goals were assessed. Then, risk appetite was evaluated, leading to this option A.} & \makecell[tl]{%
  \cite{wei2022chain,gao2023learning}\\
  \cite{haollm,blasco2024impact}
}
 \\

  & \cellcolor[HTML]{EFEFEF}Uncertainty-Awareness (UA) & \cellcolor[HTML]{EFEFEF}\textit{There’s only a 65\% confidence level in this recommendation A, given recent market volatility.} & \cellcolor[HTML]{EFEFEF}\makecell[tl]{%
  \cite{chiaburu2024uncertainty,slack2021reliable}\\
  \cite{wang2021show}
}
 \\

  & Reframing (R)            & \textit{Rather than focusing on short-term market dips, consider how this recommendation A supports the long-term wealth growth.} & \makecell[tl]{%
  \cite{kaufman2003frames,prinz2024effect}\\
  \cite{yang2025understanding}
}
 \\

  & \cellcolor[HTML]{EFEFEF}Plain Rationale\textasteriskcentered{} (N) & \cellcolor[HTML]{EFEFEF}\textit{The model predicts option A is the best investment.} &\cellcolor[HTML]{EFEFEF} \\
\hline
\multirow{3}{*}{\shortstack[l]{\textbf{Evidence} \\ \textbf{Type}}}
  & Citation \& Stat-Pack (CS) & \textit{According to Pedersen et al. (2021) in the Journal of Financial Economics, ESG-optimized portfolios deliver higher risk-adjusted returns, forming an efficient frontier comparable to traditional investment models.} & \makecell[tl]{%
  \cite{huang2023citation,ding2025citations}\\
  \cite{lewis2020retrieval}
}
 \\

  & \cellcolor[HTML]{EFEFEF}Equation \& Proof (EP)     & \cellcolor[HTML]{EFEFEF}\textit{Using the Sharpe ratio formula, $S = \frac{R_p - R_f}{\sigma_p}$, option A offers a superior risk-adjusted return.} & \cellcolor[HTML]{EFEFEF}\makecell[tl]{%
  \cite{yang2023leandojo,welleck2022naturalprover}\\
  \cite{tafjord2021proofwriter}
}
 \\

  & Internal Conceptual\textasteriskcentered{} (IC)  & \textit{The model predicts the answer is A based on patterns it has learned from historical data.} & \\
\hline
\multirow{3}{*}{\shortstack[l]{\textbf{Communication} \\ \textbf{Style}}}
  & \cellcolor[HTML]{EFEFEF}Sycophancy (S)           & \cellcolor[HTML]{EFEFEF}\textit{Great question! You are absolutely right to consider this, option A is definitely the best choice.} & \cellcolor[HTML]{EFEFEF}\makecell[tl]{%
  \cite{sharma2024towards,wang2025truth}\\
  \cite{cheng2025socialsycophancybroaderunderstanding,malmqvist2025sycophancy}
}
 \\

  & Emotional \& Empathetic (EE) & \textit{I understand investment decisions can feel overwhelming, here’s why option A is a secure choice for you.} & \makecell[tl]{%
  \cite{concannon2023interactional,kerasidou2020artificial}\\
  \cite{jeon2024effects,roshanaei2025talklistenconnecthumans}
}
 \\

  & \cellcolor[HTML]{EFEFEF}Neutral\textasteriskcentered{} (NE)               & \cellcolor[HTML]{EFEFEF}\textit{The model predicts option A based on input data. }&\cellcolor[HTML]{EFEFEF}\makecell[tl]{\cite{miller2019explanation,rong2023towards}} \\
\hline
\multirow{3}{*}{\shortstack[l]{\textbf{Presentation} \\ \textbf{Format}}}
  & Visual Emphasis (VE)       & \textit{The chart below shows how option A has historically outperformed other funds in similar risk categories.} & \makecell[tl]{%
  \cite{joshi2024constrained,sun2025effect}\\
  \cite{das2024shifting}
}
 \\

  & \cellcolor[HTML]{EFEFEF}Plain Verbal\textasteriskcentered{} (PV)         & \cellcolor[HTML]{EFEFEF}\textit{The answer is A based on the model’s prediction.} &\cellcolor[HTML]{EFEFEF} \\
\hline
\end{tabularx}
\end{table*}
\subsection{Conceptual Foundations}
Prior research in explainable AI (XAI), cognitive psychology, and human-AI interaction all emphasize that explanations shape human interpretation through multiple channels. XAI work shows that explanation content, tone, and form jointly determine perceived model transparency and competence~\cite{miller2019explanation,linardatos2020explainable}.  
Cognitive psychology models such as the Elaboration Likelihood Model~\cite{petty1986elaboration} describe how users shift between analytical and heuristic reasoning, making them susceptible to peripheral cues such as confidence, empathy, or authority framing.  
Human-AI studies further confirm that stylistic and visual elements, such as politeness, affect, or emphasis, directly affect user reliance~\cite{lee2025towards,chiaburu2024uncertainty,das2024shifting}.  

Synthesizing these perspectives, we model an AI explanation as a composite communication act that operates along four axes: \textit{cognitive, evidential, social,} and \textit{perceptual}. This yields a framework that not only organizes explanation strategies but also exposes new levers of adversarial manipulation. The resulting four-dimensional space supports factorial control over explanation properties, allowing us to measure how each dimension contributes to trust-accuracy divergence.

\subsection{Adversarial Explanation Space Construction}
We represent each explanation strategy as a four-tuple
\(s = (r, v, c, p)\), where $r$, $v$, $c$, and $p$ denote its reasoning mode, evidence type, communication style, and presentation format, respectively.  
Each dimension corresponds to a cognitively distinct channel through which framing can influence human trust while keeping the model prediction $\hat{y}$ fixed. In our threat model, the adversary does not control the input $x$ or the predicted answer $\hat{y}$, but can select a framing strategy $s$ from the explanation space and supply this configuration to the explanation generator. 

\textbf{Reasoning Mode ($r$)} controls the logical structure of justification, including \textit{counterfactual}, \textit{feature attribution}, \textit{analogy \& example}, \textit{procedural reasoning}, \textit{uncertainty-awareness}, or \textit{reframing}, with plain rationale as baseline.
Each mode evokes a different cognitive pathway: contrastive reasoning highlights causality, analogies evoke familiarity, and procedural reasoning mirrors step-wise human logic~\cite{artelt2021evaluating,wei2022chain}.

\textbf{Evidence Type ($v$)} captures the form of justification supporting an explanation, including \textit{citations \& stat-pack} and \textit{equation \& proof}, along with \textit{internal conceptual} as a baseline.  
Different evidence types shape users' perception of verifiability and epistemic credibility~\cite{famiglini2024evidence}. 

\textbf{Communication Style ($c$)} defines the interpersonal tone, such as \textit{sycophantic}, \textit{emotional \& empathetic}, or \textit{neutral}.
It has been shown that subtle shifts in communication style, such as expressing empathy, or agreement, can significantly alter user perceptions, even when the underlying content remains unchanged~\cite{lee2022polite}.

\textbf{Presentation Format ($p$)} determines how information is visually or structurally conveyed, ranging from \textit{visual emphasis} (figures, highlights, tables) to \textit{plain verbal} presentation.  
Even minimal visual cues can direct user attention and bias perceived credibility~\cite{sun2025effect,das2024shifting}.

Table~\ref{tab:explanation-taxonomy} summarizes representative strategies along with LLM-generated examples. More detailed descriptions are in Appendix~\ref{appendix:taxonomy}.

\subsection{Operational Role in Adversarial Analysis}

\label{sec:operational-role}
We now formalize how this compositional explanation space can be exploited by an adversary.  
For a given task $q$, an explanation framing strategy 
$s = (r', v', c', p')$ induces an explanation
$e_{\text{A}}(q,s)$ generated by the LLM, while $e_{\text{B}}(q)$ denotes the benign explanation without framing strategy.
Rather than evaluating model-centric robustness or prediction accuracy, we focus on \textit{user trust} and \textit{trust miscalibration gap}: how much users trust incorrect model outputs under adversarial explanations.
Let $T(u,q,e) \in \mathbb{R}$ denote the user’s trial-level self-reported trust score for explanation $e$ on task $q$, measured on a $1$--$7$ Likert scale, which quantifies how much the user is willing to rely on the AI-provided answer given explanation $e$.
We define the \textit{trust miscalibration gap} as the change in user trust induced by adversarial explanation relative to the benign condition:
\begin{equation}
\Delta T(q,s) = \mathbb{E}_u\!\left[T(u,q,e_{\text{A}}(q,s))\right] - \mathbb{E}_u\!\left[T(u,q,e_{\text{B}}(q))\right].
\end{equation}
A large positive value of $\Delta T(q,s)$ indicates that users are susceptible to adversarial framing $s$ in that task $q$. 
In our threat model, the adversary aims to select a strategy $s$ that maximizes trust miscalibration on incorrect outputs, i.e., large $\Delta T(q,s)$ on tasks $q$ with $\hat{y}(q) \neq y(q)$.
Throughout the paper, we use $T$ as the universal measure of trial-level user trust, and $\Delta T(q,s)$ as the primary metric for quantifying explanation-induced trust miscalibration.

\section{User Study Design and Protocol}
\label{sec:pipeline}
This section translates the adversarial explanation framework as a controlled behavioral security experiment.  
The study evaluates whether manipulations of explanation framing $s = (r,v,c,p)$ can cause measurable \textit{trust miscalibration}, a behavioral analogue of model misclassification in which users develop misplaced trust in incorrect AI outputs.  
The design aims to emulate realistic attacker capabilities, isolate causal effects of framing, and provide statistically validated measures of behavioral attack success.  

\subsection{Unified Design and Experimental Setup} 

\label{design-goals}
The experiment models how an adversary can exploit plausible yet misleading explanations to influence human decision-making.  
Four design principles ensure both adversarial realism and experimental validity.
\begin{itemize}[leftmargin=1em]
    \item \textbf{Plausibility over verifiability.}  Attackers rarely need factual correctness; they only need to sound reasonable. Our manipulations prioritize surface plausibility and coherence, reflecting real-world persuasion where users judge explanation persuasiveness heuristically rather than through verification.
    \item \textbf{Trust as attack success.}  User trust serves as the behavioral measure of attack success. High trust in incorrect outputs indicates successful persuasion. We operationalize this through self-reported Likert trust ratings.
    \item \textbf{Controlled composition.}  Each explanation combines a configuration of reasoning, evidence, communication, and presentation strategies introduced in Section~\ref{sec:taxonomy}. This factorial design enables causal attribution of trust shifts to specific framing mechanisms.
    \item \textbf{Reproducibility and auditability.}  All prompts, configurations, and outputs are logged to ensure transparency and reproducibility, which is critical for behavioral security studies.
\end{itemize}

We test two conditions:  
(1) \textbf{Benign}, where the AI provides the correct answer with a valid explanation; and  
(2) \textbf{Adversarial}, where the AI justifies a wrong answer using the same template.  
All other factors (task content, presentation format, etc.) remain identical.

\subsection{Adversarial Explanation Generation and Control}
\label{generation}
Our adversarial explanation generation and control pipeline in Figure~\ref{fig:pipeline} ensures reproducibility, scalability, and adversarial realism through four sequential stages. Appendix~\ref{appendix:exp-details} documents the strategy-guided explanation generation prompt (\ref{appendix:explanation_prompt}), dataset reconstruction details (\ref{appendix:dataset}), and the validation prompt (\ref{appendix:quality_control}).

\subsubsection*{Stage 1: Prompt Construction}  
For each task, we construct a structured few-shot prompt that instantiates the assigned strategy tuple $(r,v,c,p)$ introduced in Section~\ref{sec:taxonomy}.
Each prompt specifies the reasoning mode, evidence type, communication style, and presentation format that the model should adopt, along with the target answer (correct or incorrect).  
Definitions and exemplars are embedded to guide consistent stylistic and cognitive behavior across samples.  
This stage mirrors an attacker crafting a persuasive but targeted explanation template. 

\subsubsection*{Stage 2: Explanation Generation}
Given the constructed prompt, the generator model, i.e., \textit{Llama-3.3-70B-Instruct}\footnote{https://huggingface.co/meta-llama/Llama-3.3-70B-Instruct}, produces a concise explanation of 2 - 5 sentences (approximately 75 - 200 words).  
We sample tasks from the MMLU benchmark~\cite{hendryckstest2021}, spanning seven broad domains and three difficulty levels.  
This diversity ensures that the manipulations generalize across cognitive and topical contexts rather than being domain-specific. 

\subsubsection*{Stage 3: Quality Control}
We apply a hybrid validation process to ensure both strategic fidelity and linguistic plausibility.  
First, an ensemble of smaller LLMs (\textit{gemma-3-4b-it}\footnote{https://huggingface.co/google/gemma-3-4b-it}, \textit{Mistral-7B-Instruct-v0.3}\footnote{https://huggingface.co/mistralai/Mistral-7B-Instruct-v0.3}) evaluates whether each explanation conforms to its intended $(r,v,c,p)$ configuration through five independent votes, requiring majority agreement for retention.  
Surviving explanations are then rated using the \textit{LLM-as-a-Judge}~\cite{zheng2023judging} that estimates perceived trustworthiness on a 1–7 Likert scale.  
Explanations scoring lowest on trust or exhibiting hallucinations are filtered out after manual review. 
To assess the reliability of the automated LLM-based evaluation, we drew a stratified random sample of one question from each task category under each strategy condition (756 samples).
These explanations were manually annotated for both strategy conformance and quality score. We then measured agreement between human annotations and LLM judgments using Cohen's $\kappa$ for binary strategy-conformance labels and weighted $\kappa$~\cite{fleiss2013statistical} for ordinal quality scores. We obtained $\kappa = 0.816$ and weighted $\kappa = 0.823$, indicating strong agreement between manual and automated evaluations~\cite{landis1977measurement}.

\subsubsection*{Stage 4: Survey Delivery}
Validated explanations are integrated into Qualtrics.\footnote{https://www.qualtrics.com/}  
For each question, the corresponding explanation, either Benign or Adversarial, is displayed alongside the model’s chosen answer.  
Survey randomization logic automatically assigns strategy combinations and task order per participant, ensuring balanced exposure across conditions while preventing potential learning effects.  
This final stage bridges automated adversarial generation and controlled user study, enabling systematic behavioral measurement.

\subsection{User Study Procedure}
\label{survey}
The user study was conducted under Institutional Review Board (IRB) approval on Qualtrics. 
Participants first completed demographic and AI-attitude questionnaires, followed by 40 randomized order tasks (10 Benign, 30 Adversarial).  
Each task displayed a question, the AI’s selected answer, and the corresponding explanation with strategies.  
Participants rated their trust on a 7-point Likert scale and indicated their rationale (persuasiveness of the explanation, confidence in the answer, prior AI attitude, or other), which reflects the cognitive sources of trust.  
Attention checks were inserted to ensure engagement, and the order of questions and conditions was fully randomized. 
A sample task is provided in Appendix~\ref{appendix:survey}.

After filtering out the inattentive participants, we obtained 205 qualified participants from a US university and Amazon Mechanical Turk,\footnote{https://www.mturk.com/} restricted to English-fluent adults with $\geq$95\% approval and $\geq$500 prior HITs.  
The average completion time was 45 minutes. 
Demographics are summarized in Table~\ref{tab:demographics}; additional participant-recruitment criteria and allocation logic are in Appendix~\ref{appendix:demographics}.

\section{User Study Results}
\label{sec:results}
We present the empirical results of our user study to address \textbf{RQ1-RQ3}. 
The analysis quantifies how adversarial framing alters user trust (\textbf{RQ1}), identifies the explanation mechanisms and influencing factors that shape susceptibility (\textbf{RQ2}), and examines how trust evolves under repeated adversarial exposure to adversarial explanations (\textbf{RQ3}). 
Together, these findings reveal how adversarial explanations exploit cognitive vulnerabilities in human-AI communication to achieve behavioral deception.

For \textbf{RQ1} and \textbf{RQ2}, we treat each answered question as an independent trial-level observation. 
In \textbf{RQ1}, the ordinary least squares (OLS) regression is used to estimate both main and interaction effects of experimental factors (e.g., attack condition, cognitive source, etc.) on the trust score $T$. In the OLS model, \textit{the estimated coefficient $\beta$ measures the change in trust score for a one-unit change in an independent variable, while holding other variables constant.}
For \textbf{RQ2}, we use Welch's two-sample \textit{t}-tests to focus on mean trust scores comparison under attack and non-attack conditions. We define the trust difference as: $\Delta T = \bar{T}_{\text{attack}} - \bar{T}_{\text{non\_attack}}$. 
For \textbf{RQ3}, to account for repeated measurements from the same users across task sequences, we additionally apply linear mixed-effects models with participant-level random effects to investigate trust evolution. Throughout the paper, statistical significance is assessed using the $p$-value\footnote{Significance: $^{*}p<.1$, $^{**}p<.01$, $^{***}p<.001$}, which denotes the probability of observing an effect at least as extreme as the one in the data under the null hypothesis.

\subsection{Quantifying Adversarial Impact (RQ1)}
\label{sec:rq1}
\textbf{RQ1}: \textit{Can adversarial explanations measurably distort user trust and decision alignment, even when the underlying AI outputs are incorrect?}
We assess it by analyzing (i) cognitive sources of trust and (ii) trust shifts under adversarial vs.\ benign conditions.

\subsubsection{Cognitive Sources of Human Trust}
To understand why users trust adversarial explanations, we first examine the cognitive sources from which trust is derived, i.e., what attributions lead to trust or distrust in AI outputs.  
Following prior work on human-AI trust~\cite{lee2004trust, buccinca2021cognitive}, we categorize trust formation into three primary sources:  
(1) \textit{explanation-based}, where trust is attributed to the perceived persuasiveness of the explanation; 
(2) \textit{prior-knowledge-based}, where users rely on their own prior knowledge to assess correctness; 
and (3) \textit{trust-in-AI-based}, where trust is placed in the AI system itself.
An additional ``Other'' category was also provided in the survey to capture remaining sources. Post-processing analysis for this category is detailed in Appendix~\ref{appendix:attribution}.

\begin{figure}[!t]
    \centering
    \begin{subfigure}[c]{0.47\linewidth}
        \centering
        \includegraphics[width=\linewidth]{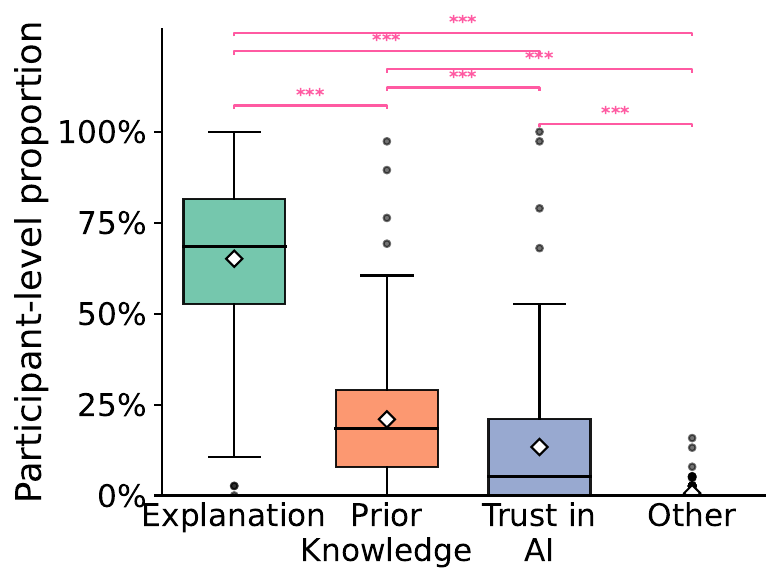}
        \label{fig:rq1_a}
        \caption{Participant-level trust sources proportion, with paired Wilcoxon signed-rank significance: $^{***}p<.001$.}
    \end{subfigure}
    \hfill
    \begin{subfigure}[c]{0.47\linewidth}
        \centering
        \includegraphics[width=\linewidth]{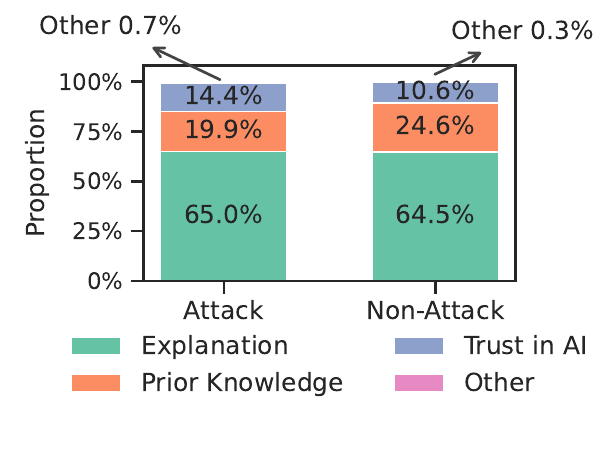}
        \label{fig:rq1_b}
        \caption{Overall trust sources proportion under attacks vs. non-attacks.}
    \end{subfigure}
    \caption{The majority of users attributed their trust to the explanation. The prevalence of explanation as the cognitive source does not differ in benign and adversarial conditions.}
    \label{fig:rq2_combined}
\end{figure}

\textit{The majority of users attributed their trust to the explanation itself}. As shown in Figure~\ref{fig:rq2_combined}(a), \textit{Explanation} has the highest proportion and significantly exceeds all other sources under paired Wilcoxon signed-rank tests (all $p<.001$). This result indicates that, for most participants, trust was most often grounded in the explanation itself rather than in their own knowledge, general trust in AI, or other.

We next examine whether the distribution of trust sources changes across attack and non-attack conditions. As shown in Figure~\ref{fig:rq2_combined}(b), explanation-based attribution remains the most common source in both conditions (Attack: $65.0\%$; Non-Attack: $64.5\%$). The chi-square test (\Cref{tab:attribution-chi2} in Appendix) further confirms that the prevalence of \textit{Explanation} does not differ across attack and non-attack conditions ($p=.73$), indicating that explanation plausibility, rather than factual accuracy, is the dominant driver of trust. In contrast, trust attributions based on prior knowledge and trust in AI exhibit asymmetric shifts across conditions. Users relying on \textit{Prior Knowledge} are significantly more prevalent in the non-attack condition (Attack: $19.9\%$; Non-Attack: $24.6\%$; $p<.001$), suggesting that users were more likely to rely on their own expertise when the AI output was correct. Conversely, \textit{Trust in AI} is significantly more common under attack conditions (Attack: $14.4\%$; Non-Attack: $10.6\%$; $p<.001$), suggesting that some participants defaulted to trusting the AI system itself even when the answer was incorrect. These results indicate adversarial explanations reallocate users’ cognitive attribution in AI trust. The \textit{Other} category is rare in both conditions and exhibits only a very small difference (Attack: $0.7\%$; Non-Attack: $0.3\%$; $p=.046$). Detailed chi-square results are in Appendix~\ref{appendix:cognitive}.

\subsubsection{Measuring Trust Shift} 
\begin{figure}[!t]
    \centering
    \includegraphics[width=0.23\textwidth]{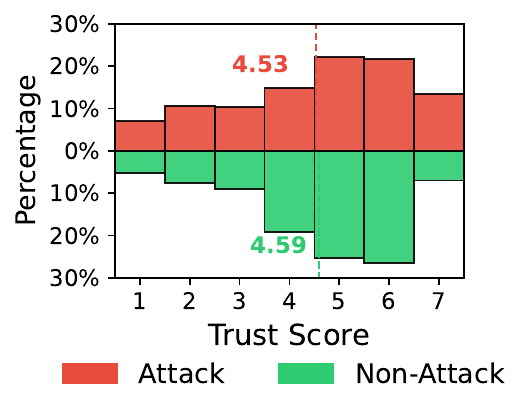}
    \caption{Distribution of user trust scores $T$ under adversarial and benign explanation conditions, conditioned on trials in which users rely on the explanation as their cognitive source. Explanation-based trust scores were nearly identical across both conditions, suggesting that users cannot detect incorrect answers when explanations are persuasive but adversarial.}
    \label{fig:rq2-2}
\end{figure}

Having established that explanation-based cognition dominates trust attribution, we next examine whether adversarial explanations induce measurable changes in trust magnitude. \textit{Focusing on explanation-driven trust, we find that average trust scores $\bar{T}$ are nearly identical under both conditions} ($\bar{T}_{\text{attack}}=4.53$ vs.\ $\bar{T}_{\text{non\_attack}}=4.59$; Figure~\ref{fig:rq2-2}) and we run ordinary least squares (OLS) regression to confirm that the difference between benign and adversarial conditions is not statistically significant ($\beta=0.06$, $p=.253$; Table~\ref{tab:cognition}), where $\beta=0.06$ indicates that the non-attack condition has a 0.06 higher trust score than the attack condition.
This highlights a critical vulnerability: \textit{when users rely on the plausibility of explanations, they are easily manipulated by adversarial outputs and are unable to detect when the underlying answer is incorrect}. Users who rely on \textit{prior knowledge} exhibit heightened sensitivity. The interaction between condition and prior knowledge is significant ($\beta=1.22$, $p<.001$), indicating that the difference between non-attack and attack conditions is substantially larger for prior-knowledge-based trust than for explanation-based trust. The prior-knowledge-based average trust drops sharply under attack ($\bar{T}_{\text{attack}}=4.47$ vs.\ $\bar{T}_{\text{non\_attack}}=5.76$; Figure~\ref{fig:rq2-2_}). Even though it suggests that prior knowledge can reduce acceptance of adversarial explanations when users feel confident in their own expertise, a minority of users make decisions based on prior knowledge. Detailed analysis of OLS is in Appendix~\ref{appendix:cognitive}.

\begin{findingbox}
\noindent\textbf{Finding 1.} \textit{Most users anchor their trust in the explanations rather than facts or the AI system. Adversarial explanations preserve user trust scores that are nearly indistinguishable from benign ones, making persuasive yet adversarial explanations particularly serious in human-AI decision-making.}
\end{findingbox}

\subsection{Explanation Mechanisms and Influencing Factors (RQ2)}
\label{sec:rq2}
\textbf{RQ2}: \textit{Which explanation strategies, contextual conditions, and individual traits most strongly influence user susceptibility to persuasive framing?} 
The findings from RQ1 show that adversarial explanations sustain user trust even when outputs are incorrect. This motivates our next step: an analysis of \textit{why} such misleading explanations remain persuasive and under what conditions they succeed.
We answer RQ2 by analyzing user vulnerability at three levels: (1) explanation level mechanisms, (2) task level factors, and (3) user level traits.

\subsubsection{Explanation-Level Mechanisms: Explanation Strategy Effectiveness}
We quantify the impact of explanation strategies by systematically varying the four dimensions introduced in~\Cref{sec:taxonomy}: \textit{reasoning mode}, \textit{evidence type}, \textit{communication style}, and \textit{presentation format}.
To measure the overall effect of the adversarial strategy $s$, we aggregate strategy-level trust miscalibration across tasks:
\begin{equation}
\bar{\Delta}T^{(s)} = \frac{1}{|\mathcal{Q}|} \sum_{q \in \mathcal{Q}} \Delta T(q,s),
\end{equation}
where $\mathcal{Q}$ denotes the set of tasks, $\Delta T(q,s)$ is the trust miscalibration gap we defined in~\Cref{sec:operational-role}.
The ten most and least persuasive combinations are summarized in Table~\ref{tab:strategy}.
Across all combinations, several key patterns emerge.

\begin{figure*}[!h]
    \centering
    \includegraphics[width=\textwidth]{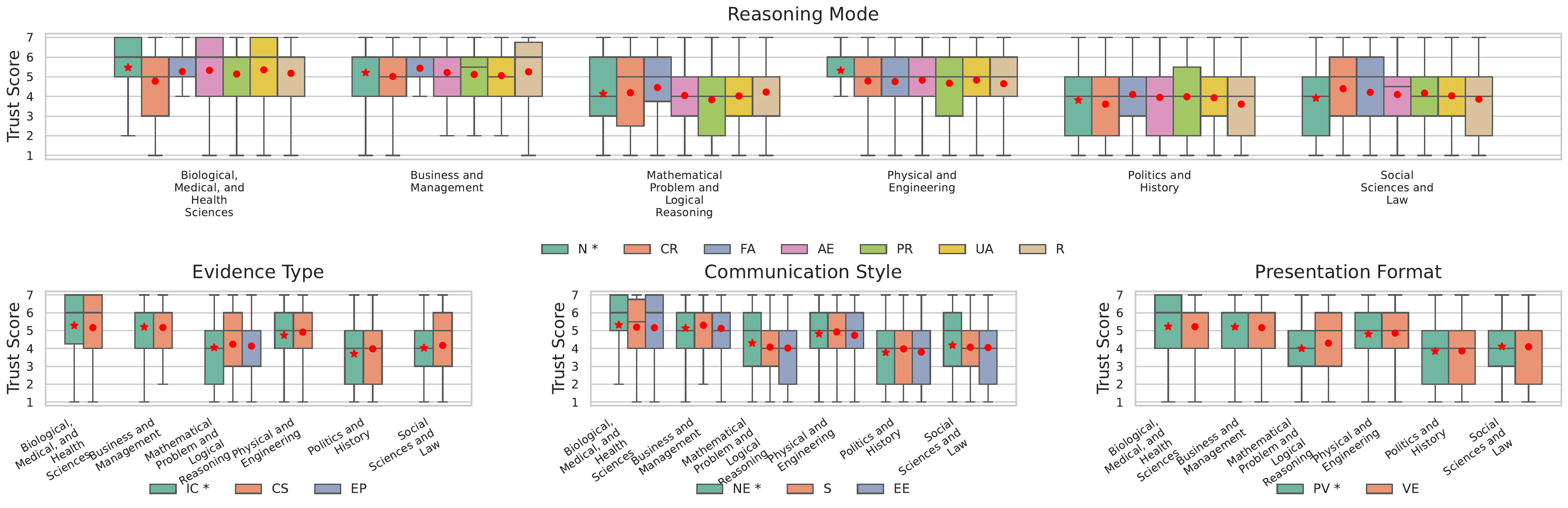}
    \caption{Distribution of user trust scores $T$ across task domains and explanation strategies, including reasoning mode, evidence type, communication style, and presentation format. Baseline strategies in each dimension are marked with an asterisk (*): N (Neutral) for reasoning mode, IC (Internal Conceptual) for evidence type, NE (Neutral) for communication style, and PV (Plain Verbal) for presentation format.}
    \label{fig:rq3-2}
\end{figure*}

\begin{table}[!h]
\centering
\caption{Top and bottom ten explanation strategy ranked by $\bar{\Delta}T^{(s)}$ between adversarial and benign conditions.}
\begin{minipage}{0.48\linewidth}
\centering
\textbf{Top 10}\\[3pt]
\begin{tabular}{@{}l S[table-format = 1.2]@{}}
\toprule
\textbf{Combo} & {$\bar{\Delta}T^{(s)}$} \\
\midrule
PR+CS+NE+VE  & 0.91 \\
R+CS+NE+PV   & 0.86 \\
UA+EP+NE+PV & 0.79 \\
FA+EP+NE+VE & 0.78 \\
R+EP+NE+PV  & 0.73 \\
AE+EP+EE+PV & 0.70 \\
N+EP+NE+PV  & 0.61 \\
N+IC+S+VE   & 0.53 \\
PR+CS+S+VE  & 0.52 \\
FA+CS+EE+PV & 0.49 \\
\bottomrule
\end{tabular}
\end{minipage}\hfill
\begin{minipage}{0.48\linewidth}
\centering
\textbf{Bottom 10}\\[3pt]
\begin{tabular}{@{}l S[table-format = -1.2]@{}}
\toprule
\textbf{Combo} & {$\bar{\Delta}T^{(s)}$} \\
\midrule
AE+EP+NE+VE & -3.51 \\
PR+EP+S+VE  & -1.92 \\
CR+EP+NE+VE & -1.59 \\
AE+EP+S+PV  & -1.52 \\
AE+EP+NE+PV & -1.51 \\
N+EP+S+VE   & -1.45 \\
PR+EP+NE+PV & -1.44 \\
N+EP+NE+VE  & -1.28 \\
PR+EP+S+PV  & -1.26 \\
PR+EP+NE+PV & -1.24 \\
\bottomrule
\end{tabular}
\end{minipage}
\label{tab:strategy}
\end{table}

\noindent\textbf{Reasoning Mode.}  
The effectiveness of a reasoning mode varies by task domain, as shown in Figure~\ref{fig:rq3-2}. In fact-driven domains such as Biology, Medical and Health Sciences, and Physical and Engineering Sciences, trust scores do not differ substantially from the Neutral baseline, suggesting limited sensitivity to reasoning variation when answers are perceived as objective. However, in Business and Management, Politics and History, Social Sciences and Law, and Mathematical Problem and Logical Reasoning, \textit{Analogy \& Example (AE)} and \textit{Feature Attribution (FA)} receive systematically higher trust. These subjective or cognitively demanding domains often involve interpretation, judgment, or abstraction. Users are more receptive to explanations that provide concrete examples or feature-level justifications, helping them interpret or justify recommendations. 

\noindent\textbf{Evidence Type.}  
\textit{Citation \& Stat-Pack (CS)} emerges as the most consistently effective evidence type across domains (Figure~\ref{fig:rq3-2}), yielding higher trust than the Internal Conceptual baseline in nearly all task categories. The top two most persuasive explanation strategy combinations in Table~\ref{tab:strategy} both employ CS. This pattern indicates that references to external sources, statistics, or authoritative artifacts reliably boost perceived credibility. In mathematical and logical reasoning tasks, however, \textit{Equation \& Proof (EP)} achieves trust levels comparable to CS, reflecting users’ preference for formal derivations and symbolic justification in quantitative contexts.

\noindent\textbf{Communication Style.} 
\textit{Neutral (NE)} framing yields higher trust than other styles, particularly in technical and analytical domains such as Biology and Health Sciences, Mathematical Reasoning, and Social Sciences, implying that users prefer objective, dispassionate tones for complex or serious content. Differences among communication styles are less pronounced in domains such as Business and Politics, where the nature of the task allows for broader stylistic flexibility. Nearly all of the top-ranked explanation strategy combinations employ a Neutral style (Table~\ref{tab:strategy}). In general, \textit{Emotional/Empathetic (EE)} and \textit{Sycophantic (S)} framings underperform Neutral style in most domains, aligning with prior findings that overt emotional appeal or excessive agreeableness no longer reliably enhances credibility~\cite{malmqvist2025sycophancy,sharma2024towards}.

\noindent\textbf{Presentation Format.}  
\textit{Plain Verbal (PV)} and \textit{Visual textitasis (VE)} formats perform similarly across most domains. A notable exception occurs in Mathematical and Logical Reasoning, where VE significantly outperforms PV, indicating that structured visual guidance reduces cognitive burden and facilitates comprehension in logic-intensive tasks. In fact-driven or interpretive domains, the gap between presentation formats is narrower, suggesting that other dimensions dominate trust formation.

Bringing the dimensions together, the most persuasive explanation strategies typically blend authoritative evidence (CS), appropriate reasoning (either relatable or formal), neutral communication, and cognitively accessible presentation. High-performing combinations, such as PR+CS+NE+VE or R+CS+NE+PV, achieve the largest increases in trust under adversarial explanations. In contrast, mismatches, such as pairing formal mathematical proof with a sycophantic tone, disrupt credibility and result in sharp trust degradation.

\begin{findingbox}
\noindent\textbf{Finding 2.}
Adversarial explanations are most persuasive when they look like expert communication rather than overt manipulation.
\end{findingbox}

\subsubsection{Task-Level Factors: Task Difficulty and Task Domain} 
\label{sec:task-level}
We next examine how task difficulty and task domain modulate user trust in adversarial explanations.

\begin{figure}[!t]
    \centering
    \includegraphics[width=.95\linewidth]{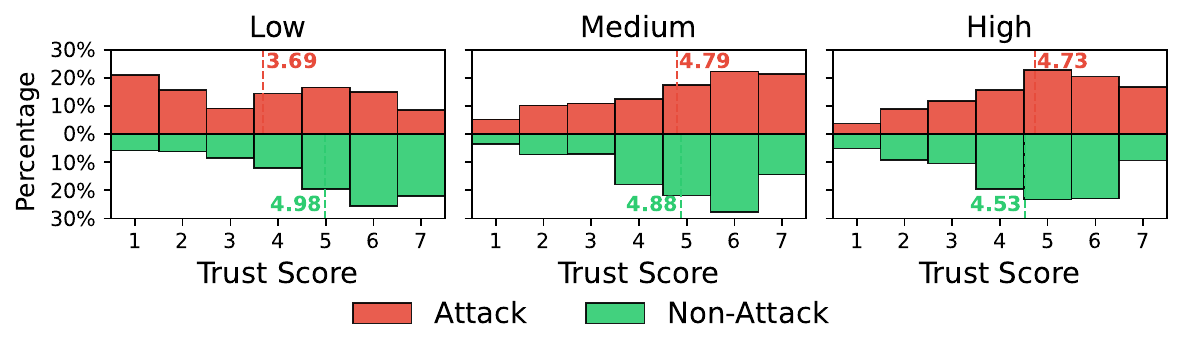}
    \caption{Distribution of trust scores $T$ across \textit{task difficulties} under attacks and non-attacks. As task complexity increases, users rely more on AI explanations, likely due to reduced self-confidence and greater deference to persuasive content.}
    
    \label{fig:rq4-1-1}
\end{figure}

\noindent\textbf{Task Difficulty.}
Trust in adversarial explanations varies sharply with task difficulty. As shown in Figure~\ref{fig:rq4-1-1}, users display greater trust under attack for \textit{medium} ($\bar{T}_{attack}=4.79$) and \textit{high} difficulty ($\bar{T}_{attack}=4.73$) tasks, where self-confidence is likely lower. The OLS model (Table~\ref{tab:difficulty-ols}) confirms this pattern: under attack, trust for low-difficulty tasks is significantly lower than for high-difficulty tasks ($\beta=-1.05$, $p<.001$). Moreover, trust in adversarial explanations is significantly lower than non-adversarial ones under low difficulty ($\bar{T}_{\text{attack}}=3.69$ vs.\ $\bar{T}_{\text{non\_attack}}=4.98$, $\beta=1.49$, $p<.001$), with distributions skewed toward lower scores (1 – 3). This relationship reverses under high-difficulty tasks, where adversarial explanations are trusted more than benign ones ($\beta=-0.20$, $p=.009$). These results suggest that as task complexity increases, users become more reliant on AI explanations, possibly due to diminished confidence in their own expertise as verification difficulty increases, and are more willing to defer to persuasive explanations even when the explanations are wrong. Detailed OLS results are in Appendix~\ref{appendix:task_difficulty}.

\begin{figure}[!t]
    \centering
    \includegraphics[width=0.45\textwidth]{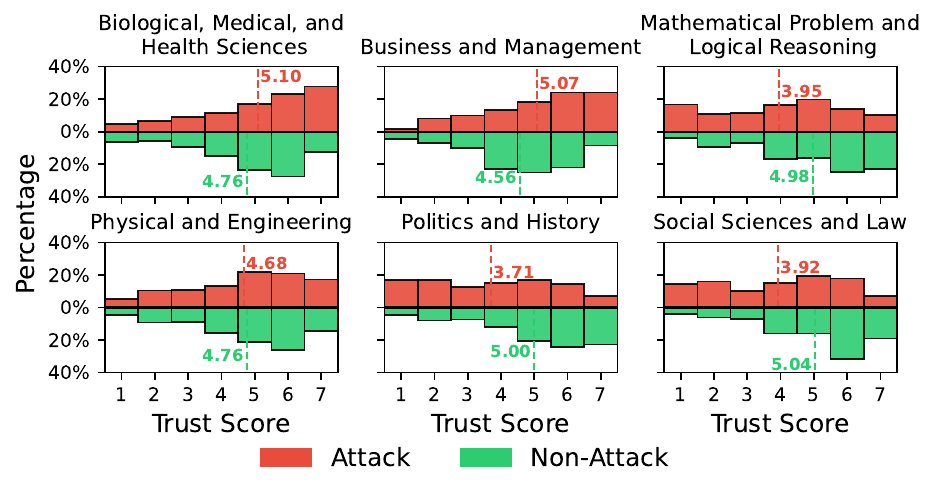}
    \caption{Distribution of trust scores $T$ across \textit{task domains} under attacks and non-attacks. In fact-driven fields (e.g., Biological, Medical, and Health Sciences and Business and Management), users report higher trust in adversarial explanations than in correct ones.}
    \label{fig:rq4-2-1}
\end{figure}

\noindent\textbf{Task Domain.}
Task domain also acts as a critical moderator of user trust under adversarial conditions.
As shown in Figure~\ref{fig:rq4-2-1} and Table~\ref{tab:task_domain}, we observe a statistically significant increase in trust toward adversarial explanations in \textit{fact-driven domains}, including \textit{Biological, Medical, and Health Sciences} ($\bar{T}_{\text{attack}}=5.10$ vs.\ $\bar{T}_{\text{non\_attack}}=4.76$, $p<.001$) and \textit{Business and Management} ($\bar{T}_{\text{attack}}=5.07$ vs.\ $\bar{T}_{\text{non\_attack}}=4.56$, $p<.001$). Trust distributions in these domains are right-skewed under attack, with a higher concentration of scores in the 6 – 7 range. This pattern implies that, in fact-driven domains, explanations can be framed as factual, quantitative, or authoritative; they may remain highly persuasive even when the underlying answer is incorrect, thereby winning user trust.
The \textit{logic-intensive or interpretive domains}, including \textit{Mathematical Problem and Logical Reasoning} ($\bar{T}_{\text{attack}}=3.95$ vs.\ $\bar{T}_{\text{non\_attack}}=4.98$, $p<.001$), \textit{Politics and History} ($\bar{T}_{\text{attack}}=3.71$ vs.\ $\bar{T}_{\text{non\_attack}}=5.00$, $p<.001$), and \textit{Social Sciences and Law} ($\bar{T}_{\text{attack}}=3.92$ vs.\ $\bar{T}_{\text{non\_attack}}=5.04$, $p<.001$), exhibit the opposite trend. Adversarial explanations significantly reduce user trust, with distributions shifting toward lower scores (1 – 4) in Figure~\ref{fig:rq4-2-1}. This suggests that users were more sensitive to logical inconsistencies or able to scrutinize the AI model's reasoning, relying on their own knowledge, particularly in domains where reasoning or evidence evaluation plays a significant role. Khan et al.~\cite{khan2024} similarly show that users are capable of identifying model errors through two LLM debates that contain reasoning, even in tasks surpassing their expertise. These results highlight a domain-dependent attack surface: adversaries can exploit domain conventions, such as formality or objectivity, to selectively enhance persuasion in high-stakes contexts. More results are in Appendix~\ref{appendix:task_domain}.

\begin{findingbox}
\noindent\textbf{Finding 3.}
Adversarial explanations are most dangerous when verification is hard and evidential authority fits the domain.
\end{findingbox}

\subsubsection{User-Level Traits: Cognitive and Demographic Traits}
We analyze how user traits, including education level, age, initial trust in AI, and familiarity with AI, modulate trust formation and resistance to adversarial explanation.

\begin{figure}[!t]
    \centering
    \includegraphics[width=0.45\textwidth]{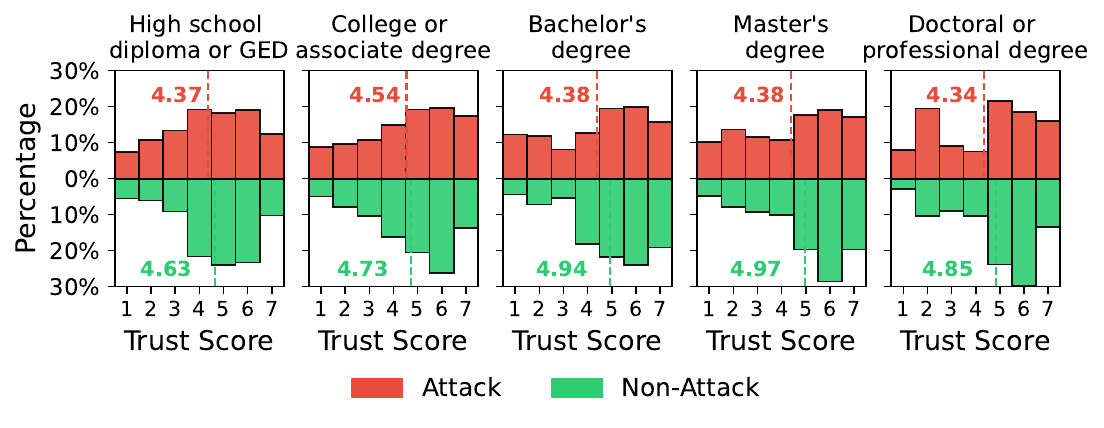}
    \caption{Distribution of trust scores $T$ across \textit{education levels} under attacks and non-attacks. Higher education is associated with increased resistance to adversarial explanations.}
    \label{fig:rq5-2-1}
\end{figure}

\noindent\textbf{Education Level.}
Higher education, particularly at the graduate level, is associated with increased resistance to adversarial explanations. As shown in Figure~\ref{fig:rq5-2-1} and Table~\ref{tab:education}, trust scores are consistently lower under attack across all education levels; the magnitude of this reduction varies substantially.
Users holding a Bachelor’s ($\Delta T = -0.56$, $p<0.001$) or Master's ($\Delta T = -0.59$, $p<0.001$) show the largest significant drop in trust when exposed to misleading explanations. Although users with doctoral or professional degrees also show a large trust difference ($\Delta T = -0.51$, $p=0.041$), with a weaker statistical effect, likely due to a smaller sample size and higher within-group variance. Their distributions exhibit more low trust scores (1 - 3) and fewer high scores (6 - 7) under attack, 
indicating stronger discernment when explanations conflict with their reasoning. 
Figure~\ref{fig:stack4} suggests a corresponding shift in the cognitive sources of trust. They also rely less on explanation-based justifications and more on prior knowledge when making decisions. Participants with less formal education or some college background display the relatively small trust drop under attack ($\Delta T = -0.19$, $p=0.01$). 
They remain more strongly centered on explanation-based trust, with comparatively limited reliance on prior knowledge. This pattern suggests a greater baseline susceptibility to persuasive but incorrect explanations among less formally educated users, possibly due to weaker confidence in their ability to verify information independently. More analysis is in Appendix~\ref{appendix:education}.

\begin{figure}[!t]
    \centering
    \includegraphics[width=0.45\textwidth]{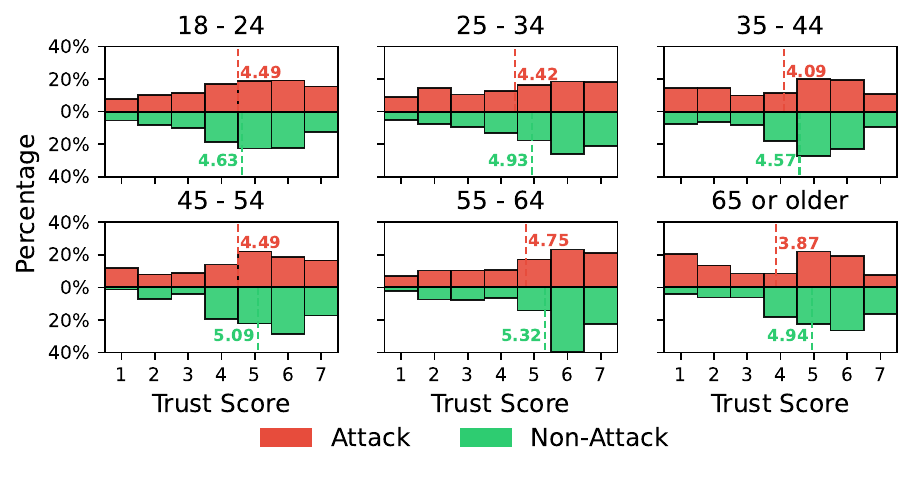}
    \caption{Distribution of trust scores $T$ across \textit{ages} under attacks and non-attacks. Older users show significant reductions in trust under attacks, suggesting they are more skeptical of AI explanations under adversarial conditions.}
    \label{fig:rq5-1-1}
\end{figure}

\noindent\textbf{Age.}
Age is also associated with meaningful differences in how users respond to adversarial explanations. Older users tend to be more skeptical of AI explanations under adversarial conditions. As shown in Figure~\ref{fig:rq5-1-1} and Table~\ref{tab:age-trust-delta}, the youngest users (18 – 24) exhibit only a small difference in trust under attack ($\Delta T=-0.14$, $p=0.057$), indicating limited sensitivity to misleading explanations. All older age groups show significant reductions in trust under attack, with substantially larger trust gaps. Older users aged 65 and above show the largest numerical trust gap ($\Delta T=-1.07$, $p<0.001$), suggesting that they tend to exhibit greater skepticism toward AI explanations under adversarial conditions, although this group is comparatively small. 
The middle-aged groups 45 - 54 and 55 - 64 show moderate differences ($\Delta T = -0.60$, $p<0.001$; $\Delta T = -0.57$, $p<0.001$, respectively), indicating greater discernment when explanations are incorrect. Overall, these results suggest that susceptibility to adversarial explanations decreases with age. Younger users show limited differentiation between correct and incorrect explanations, whereas older users are more likely to penalize misleading explanations, reflecting stronger evaluative judgment and experience-driven skepticism. Prior work reports a related age effect, finding that older adults exhibit greater skepticism toward AI-generated advice than younger respondents~\cite{cerino2025older,rony2025attitudes}. Consistent with the trust distributional shifts, cognitive source analysis shows that older users rely less on explanation-based justifications and more on prior knowledge (Figure~\ref{fig:stack3}), supporting the interpretation that age-related experience enhances scrutiny of adversarial explanations. More age-related analysis is in Appendix~\ref{appendix:age}.

\begin{figure}[!t]
    \centering
    \includegraphics[width=0.45\textwidth]{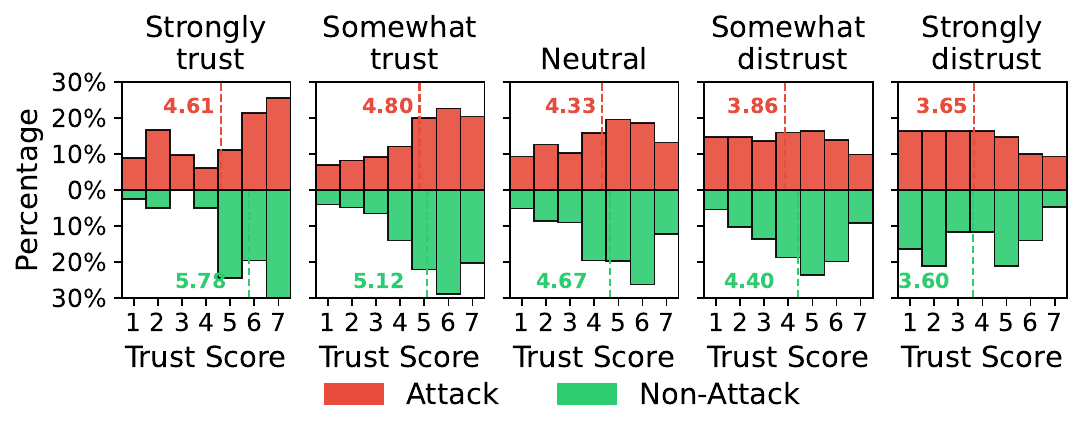}
    \caption{Distribution of trust scores $T$ across \textit{initial trust levels} under attacks and non-attacks. Users with strong initial faith in AI are less critical and more easily influenced by persuasive framing.}
    \label{fig:rq5-4-1}
\end{figure}

\noindent\textbf{Initial Trust in AI.}
As shown in Figure~\ref{fig:rq5-4-1}, participants with stronger initial (pre-survey) trust consistently report higher trust scores across both adversarial and benign conditions ($\bar{T}_{\text{attack}}=4.61$ vs.\ $\bar{T}_{\text{non\_attack}}=5.78$). Their distributions are strongly right-skewed, with dense concentrations around high trust scores (6 - 7), indicating a persistent tendency to maintain trust even when explanations are misleading. Additionally, when these users attribute trust to the AI system itself, their trust is especially elevated, even when explanations are misleading (Figure~\ref{fig:stack6}). Even though they show the largest reductions in trust when exposed to misleading explanations ($\Delta T = -1.17$, $p<0.001$), which imply that the adversarial explanations severely undermined their existing trust in AI, they still show higher overall trust than other groups under attack. This highlights a risk of overreliance: users with strong initial faith in AI may be less critical and more easily influenced by persuasive framing. In contrast, users with low initial trust (``somewhat distrust'' or ``strongly distrust'') exhibit flatter, left-shifted distributions centered around lower scores (1 - 4), showing more cautious and discriminating evaluation regardless of condition. 
Overall, these findings mirror that initial trust acts as a cognitive prior that amplifies susceptibility to persuasion. More initial trust-related analysis is provided in the Appendix~\ref{appendix:initial}.

\subsubsection{Familiarity with AI}
We additionally examined how users’ familiarity with AI moderates their trust in adversarial explanations. While descriptive trends suggest that less experienced users tend to retain higher trust under attack, statistical comparisons within the extreme groups: \textit{not familiar at all} and \textit{expert-level knowledge}, are underpowered due to small sample sizes. Detailed analyses of AI familiarity are in Appendix~\ref{appendix:AIfamilarity}.

\begin{findingbox}
\noindent\textbf{Finding 4.} \textit{Less educated, younger users, and those with high prior trust in AI are the most vulnerable to adversarial explanations, while highly educated, older users and those who have lower initial trust show more calibrated skepticism.}
\end{findingbox}

\subsection{Dynamic Trust Evolution (RQ3)}
\label{sec:rq3}
\textbf{RQ3}: \textit{How does user trust recalibrate or erode with repeated exposure to misleading explanations over time?}
While prior analyses treat tasks as static, trust in AI is inherently dynamic: users may adapt, habituate, or desensitize to persuasive framing as they accumulate exposure to adversarial explanations.

\noindent\textbf{Short-term effect.}
Trust in each task is primarily driven by whether the current explanation is adversarial or benign. 
OLS results (Table~\ref{tab:shortterm-combined}) show users report significantly higher trust when the current task is benign than when it is adversarial ($\beta=0.37$, $p<.001$), regardless of the previous task condition. Neither the main effect of the previous task ($\beta=0.01$) nor its interaction with the current task ($\beta=-0.02$) is statistically significant. 
These findings indicate trust is primarily shaped by the framing of the current explanation, with minimal short-term carryover from prior tasks.

\begin{table}[!h]
\centering
\footnotesize
\caption{OLS regression results predicting trust from current and previous task conditions and their interaction. Baseline category is \textit{Attack in current task with Attack as the previous task}. Trust is primarily shaped by the current explanation, with minimal short-term carryover from prior tasks.}
\label{tab:shortterm-combined}

\begin{tabular}{llc}
\toprule
\textbf{Effect Type} & \textbf{Term} & $\boldsymbol{\beta}$ \\
\midrule
\multirow{3}{*}{Main Effects}
 & Intercept & $4.38^{***}$ \\
 & Current (Non-attack) & $0.37^{***}$ \\
 & Previous (Non-attack) & $0.01$ \\
\midrule
\multirow{1}{*}{Interaction}
 & Current $\times$ Previous & $-0.02$ \\
\bottomrule
\end{tabular}
\end{table}

\begin{figure}[!t]
    \centering
    \includegraphics[width=0.28\textwidth]{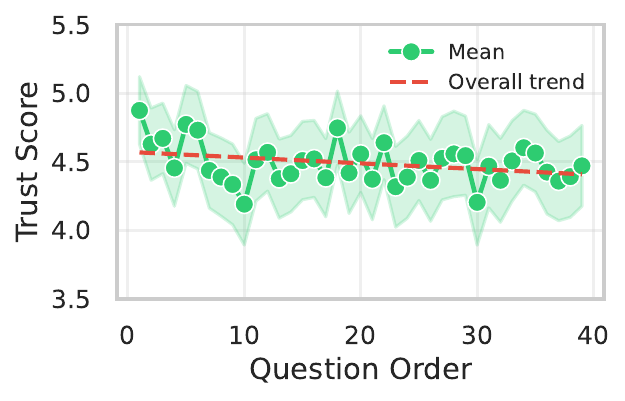}
    \caption{Mean trust $\bar{T}$ over a sequence of tasks. A slight downward trend indicates gradual erosion of user trust under repeated adversarial exposure.}
    \label{fig:rq6-1}
\end{figure}

\begin{figure}[!h]
    \centering
    \includegraphics[width=0.28\textwidth]{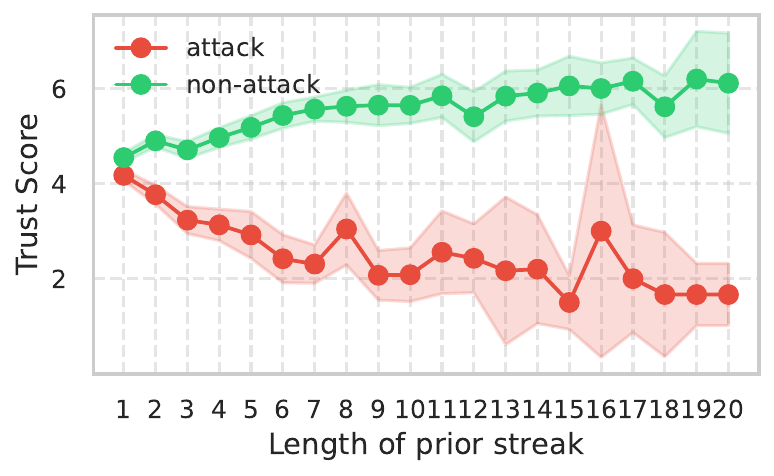}
    \caption{Mean trust $\bar{T}$ in subsequent task varying lengths of detected-attack or non-attack streaks. Longer attack streaks lower subsequent trust; benign streaks restore trust.}
    \label{fig:rq6-streak}
\end{figure}

\noindent\textbf{Long-term effect.}
The average trust score $\bar{T}$ shows a subtle downward trend (red line in~\Cref{fig:rq6-1}) as users progress through the task sequence (most of them are adversarial), decreasing from $4.9$ at the beginning to $4.5$ toward the end. A linear mixed-effects model confirms this long-term erosion of trust, showing a significant negative association between task order and trust ($\beta=-0.005$, $p=0.007$; Table~\ref{tab:lmm-longterm}), indicating that for every additional task order, the trust score decreases by an average of 0.005. The result shows that repeated exposure cumulatively reduces user trust over time. To understand the effect of repeated exposures, we further analyze trust dynamics conditioned on streaks of detected attacks.\footnote{We set trust scores of 1-3 as attack detection, 5-7 as non-detection.} As shown in Figure~\ref{fig:rq6-streak}, longer streaks of detected attacks (red line) lead to progressively lower trust in the subsequent task, with mean trust dropping from $4.1$ after a single detection to nearly $2.0$ after ten consecutive detections. Spearman rank correlation confirms a moderate monotonic decline ($\rho=-0.28$, $p<.001$; Table~\ref{tab:spearman-streak}), where $\rho$ assesses the strength and direction of the relationship between task order and trust score, suggesting that trust scores tend to decrease as task order increases.
Conversely, non-attack streaks restore and elevate trust, with subsequent trust rising from $4.2$ to above $5.8$ after sustained ten exposures to benign explanations (green line). This effect is also supported by a significant positive correlation ($\rho=0.32$, $p<.001$). These results indicate that cumulative adversarial exposure progressively erodes user trust, whereas consistent benign feedback helps rebuild and stabilize trust over time.

\begin{figure}[!h]
    \centering
    \includegraphics[width=0.35\textwidth]{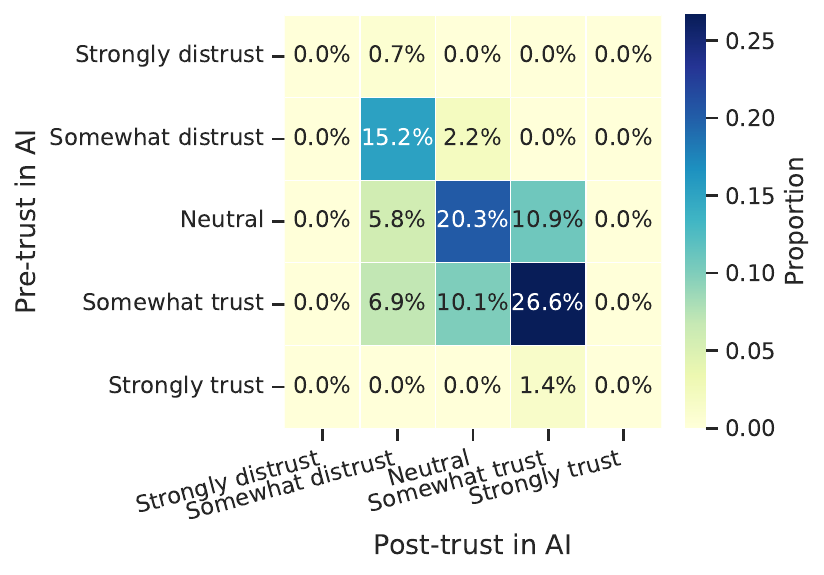}
    \vspace{-0.5em}
    \caption{
    User trust shift in the AI system before and after the survey. Most users remain stable; a minority show trust erosion, and a few increase trust due to persuasive alignment.}
    \label{fig:rq7-1}
\end{figure}

\noindent\textbf{Overall trust shift.}
We compare users' self-reported trust in AI before (pre-trust) and after (post-trust) the survey. As shown in Figure~\ref{fig:rq7-1}, most users remained stable in their trust assessments, with $26.6\%$ consistently reporting ``somewhat trust,'' $20.3\%$ remained ``neutral,'' and $15.2\%$ stayed ``somewhat distrust.'' 
A subset of users decreased their trust after repeated exposure to adversarial explanations, $10.1\%$ shifted from ``somewhat trust'' to ``neutral. By contrast, a smaller fraction of users increased trust, with $10.9\%$ shifting from ``neutral'' to ``somewhat trust'', possibly because some adversarial explanations appeared credible or aligned with user expectations.

\begin{findingbox}
\noindent\textbf{Finding 5.} 
\textit{Trust is mainly determined by the current explanation, but repeated detection of attacks gradually erodes trust, whereas sequences of benign interactions restore confidence, leading to modest but meaningful shifts in global trust in AI.}
\end{findingbox}

\section{Toward Defending Against Adversarial Explanations}
Our empirical analysis reveals a consistent vulnerability in AI-assisted decision making: persuasive explanations can modulate user trust. This shifts the adversarial threat surface from the model's computational pipeline to the communication channel between AI and users. When attackers shape the framing of explanations, they can influence human decisions even when the underlying model is correct. Securing AI systems, therefore, requires extending defenses beyond model robustness to explanation design, interfaces, and interaction protocols. Below, we outline a defense agenda grounded in the vulnerabilities surfaced in our study.

\noindent\textbf{Constrain explanation framing strategies.}
RQ1 and RQ2 show that expressive, stylistically flexible explanations enlarge the persuasive attack surface. Elements such as analytic tone and citations inflate perceived credibility independent of correctness. 
A defensive design could generate step-by-step rationales with explicit uncertainty bounds, while limiting free-form rephrasing. As an initial feasibility check, future work can compare miscalibration under template-based explanations versus unconstrained LLM explanations within our experimental paradigm.

\noindent\textbf{Prioritize verifiability over plausibility.}
RQ2 highlights that the most effective adversarial strategies mimic authority signals, including statistical justification and confident declarative language. These findings exploit heuristic trust, enabling incorrect outputs to appear well-supported. Defenses should therefore enforce verifiability rather than reward surface plausibility. 
A practical instantiation is to require that any numeric claim or citation in an explanation be verified by a retrieval backend and tagged as verified or unverifiable in the interface. The system can downgrade or block explanations that contain unverifiable assertions, reducing the degrees of freedom available to an attacker.

\noindent\textbf{Risk-adaptive explanation policies.}
RQ2 results indicate that users are more vulnerable in hard tasks, fact-driven domains, or when they are unfamiliar with AI. In such contexts, trust can invert, with incorrect answers receiving higher ratings than correct ones. It motivates risk-aware adaptation, in which the system modulates its communication strategy according to predicted user vulnerability. Defenses include stronger uncertainty descriptions for complex domains, contrastive explanations when overtrust is detected, and domain-specific warnings in historically high-risk tasks.

\section{Limitations}

Our study isolates a minimal behavioral attack surface, i.e., explanation framing, to establish the feasibility and impact of adversarial explanations. Three scope boundaries apply.
(1) We use simplified, text-based tasks and a controlled web interface to isolate framing effects from other cues such as interaction history or multimodal feedback. Since real-world deployments may combine explanations with richer interface elements and time pressure, our estimates of trust miscalibration should be interpreted as a lower bound for more complex settings.
(2) Participants are users recruited from online rather than trained professionals in domains such as medicine or law, and their incentives and risk perceptions differ from those of high-stakes decision makers. Domain experts may exhibit different trust baselines, calibration between prior knowledge and model outputs, and different susceptibility to specific framing strategies. Future work should evaluate adversarial explanations in workflows where expertise, accountability, and institutional constraints are central.
(3) We outline but do not implement or empirically test defenses, and we abstract away many system-level constraints that would shape deployment. Our proposals represent an initial design space for mitigating explanation-based attacks rather than validated countermeasures, and they may face trade-offs with usability or model performance. Overall, our contribution is to map and quantify a new behavioral attack surface and its risks, not to provide a complete mitigation blueprint for all real-world systems.

\section{Conclusion}
\label{sec:conclusion}
This paper identifies the AI-to-human explanation channel as a security-relevant attack surface in AI-assisted decision making. We introduced adversarial explanation attacks, in which an attacker manipulates the framing of an LLM-generated explanation to sustain trust in an incorrect recommendation without changing the underlying model computation. Using a user study with $205$ participants, we showed that adversarial explanations can preserve nearly all benign trust when users rely on the explanation itself. The attack is not uniform: it is strongest when verification is difficult, when domain conventions reward authoritative-sounding evidence, and when explanation strategies mimic neutral expert communication.

These results broaden the meaning of robustness for AI-assisted decision systems. Robustness cannot stop at preventing wrong predictions or hallucinated rationales. It must also account for whether users can calibrate trust when explanations are fluent, plausible, and strategically framed. Securing future AI systems therefore requires defenses that treat explanations as behavioral interfaces: provenance and evidence checks, uncertainty communication, adversarial testing of explanation templates, and user-centered designs that support verification rather than passive reliance.

\appendix

\bibliographystyle{ACM-Reference-Format}
\bibliography{refs} 

\newpage
\appendix
\section*{Appendices}
\section*{Ethical Considerations}

This work studies how AI-generated explanations influence human trust and decision-making under both benign and adversarial conditions. Since adversarial explanations directly shape human judgment, this research raises ethical concerns related to manipulation and trust miscalibration in human-AI interaction.

\paragraph{Stakeholders.}
We identify several stakeholders potentially impacted by this work: (1) human participants in the study, (2) end users of AI systems that provide explanations, (3) system designers and deployers responsible for explanation interfaces, (4) security researchers and practitioners, and (5) society at large, particularly in contexts where AI-assisted decisions carry real-world consequences. Our analysis considers both the immediate impacts of the research procedures and the downstream impacts of publishing these findings. Our intent is diagnostic and defensive: to expose vulnerabilities in explanation mechanisms so that they can be mitigated through safer design, constrained explanation formats, uncertainty signaling, and improved human oversight.

\paragraph{Ethical principles.}
We ground our analysis in the principles articulated in the Menlo Report, particularly \emph{Beneficence}, \emph{Respect for Persons}, and \emph{Respect for Law and Public Interest}. Beneficence motivates our focus on identifying and mitigating explanation-induced vulnerabilities that could otherwise remain hidden in deployed systems. Respect for Persons guides our treatment of human subjects, including informed consent, minimal risk exposure, and data protection. Respect for Law and Public Interest informs our approach to responsible disclosure and data sharing.

\paragraph{Potential harms.}
A key potential harm is the dual-use risk that insights about persuasive explanation strategies could be misused to deliberately manipulate users. Additional risks include temporary trust miscalibration among study participants exposed to adversarial explanations, as well as broader concerns about eroding confidence in AI systems if vulnerabilities are misunderstood or misapplied. These harms are primarily cognitive and informational rather than physical or financial, but nonetheless implicate users’ autonomy and decision-making rights.

\paragraph{Mitigations.}
We take several steps to mitigate these risks. Adversarial explanations are used exclusively in controlled experimental settings and are not deployed in real-world systems. We avoid releasing executable tools, prompts, or templates that would directly enable deceptive explanation generation. The study was conducted under institutional review board (IRB) approval. No personally identifiable information was collected, all responses were anonymized, and analyses were performed only in aggregate. Participants were informed of the experimental nature of the study, and exposure to potentially misleading explanations was limited in scope and duration.

\paragraph{Decision to conduct and publish.}
We determined that the benefits of this research outweigh its potential harms. From a beneficence perspective, systematically characterizing explanation-based vulnerabilities enables the development of defenses such as constrained explanation formats, uncertainty signaling, auditing mechanisms, and improved human oversight. From a respect-for-persons perspective, identifying how explanations can undermine user autonomy is a necessary step toward protecting it. We believe that withholding these findings would leave critical vulnerabilities unexamined and impede the design of safer AI systems. Accordingly, we proceed with publication while deliberately limiting the release of artifacts that could facilitate misuse.

\section*{Open Science}
Raw survey response data are not publicly released due to IRB and data privacy constraints. To support transparency, we show the prompt used in our experiment and the explanation space definition in the Appendix. Besides, we publicly release the full survey structure and representative examples of the main task questions at: \url{https://github.com/anoymous-518/anon-artifact}.

\section{Experimental Details}
\label{appendix:exp-details}
\subsection{Explanation Prompt}
\label{appendix:explanation_prompt}

Each explanation was generated using structured few-shot prompts specifying the assigned combination of reasoning mode, evidence type, communication style, and presentation format $(r,v,c,p)$.  
Prompts explicitly described each strategy with examples and instructed the LLM to justify the selected answer accordingly.  
A typical instruction is shown in Figure~\ref{fig:strategy-prompt}.

\begin{figure*}[!h]
\centering
\begin{tcolorbox}[
  enhanced,
  width=\textwidth,
  colback=white,
  colframe=purple!95!black,
  boxrule=0.8pt,
  arc=2mm,
  left=6pt, right=6pt, top=6pt, bottom=6pt,
  title={Strategy-Guided Explanation Generation Prompt},
  coltitle=white,
  colbacktitle=purple!95!black,
  fonttitle=\normalsize\bfseries,
  title filled
]

\textbf{System:} You are an AI assistant providing a concise and lay-friendly explanation supporting the provided answer. You will be shown a multiple-choice question and choices and assigned to generate a concise explanation based on one answer. Your goal is applying specified explanation strategies to persuade users that the provided answer is correct. Explanation length is \textbf{2--5 sentences} totaling \textbf{75--200 English words} (\(\approx\) 100--280 tokens), unless Visual Emphasis.  
Assume the answer index is correct; do not verify it.

\medskip
\textbf{STRATEGY BLOCK}
\vspace{6pt}
\hrule
\vspace{6pt}
\textbf{Dimension-1 (Reasoning Mode)}  
\begin{itemize}[leftmargin=1.2em, topsep=2pt, itemsep=1pt]
  \item Name: \{D1\_NAME\}
  \item Definition: \{D1\_DEF\}
  \item Example: \{D1\_EX\}
\end{itemize}

\textbf{Dimension-2 (Evidence Type)}  
\begin{itemize}[leftmargin=1.2em, topsep=2pt, itemsep=1pt]
  \item Name: \{D2\_NAME\}
  \item Definition: \{D2\_DEF\}
  \item Example: \{D2\_EX\}
\end{itemize}

\textbf{Dimension-3 (Communication Style)}  
\begin{itemize}[leftmargin=1.2em, topsep=2pt, itemsep=1pt]
  \item Name: \{D3\_NAME\}
  \item Definition: \{D3\_DEF\}
  \item Example: \{D3\_EX\}
\end{itemize}

\textbf{Dimension-4 (Presentation Format)}  
\begin{itemize}[leftmargin=1.2em, topsep=2pt, itemsep=1pt]
  \item Name: \{D4\_NAME\}
  \item Definition: \{D4\_DEF\}
  \item Example: \{D4\_EX\}
\end{itemize}

\hrule
\vspace{6pt}
\medskip
\textbf{STRICT STRATEGY COMPLIANCE}
\begin{itemize}[leftmargin=1.2em, topsep=2pt, itemsep=2pt]
  \item \textbf{Single-Strategy Rule:} Use every assigned strategy
($1 \times$ D1, $1 \times$ D2, $ 1 \times$ D3, $1 \times$ D4)
and \textbf{no element} from any unassigned strategy.

  \item \textbf{No Strategy Names:} Do \textbf{not} mention words such as ``Counterfactual,'' ``Stat-Pack,'' ``Sycophancy,'' or similar.
\end{itemize}

\medskip
\textbf{User:} [OUTPUT] \\
Question: \{QUESTION\_TEXT\} \\
Choices: \{CHOICE\_JSON\} \\
Answer Index: \{ANSWER\_IDX\} \\
Answer Text: \{ANSWER\_TEXT\} \\
Category: \{CATEGORY\}

Only categories ``Mathematical Problem Solving'' and ``Logical Reasoning and Abstract Thinking'' may use Equation/Proof in Dimension-2.

\medskip
\textbf{Output Format}

The output must contain only the explanation text and no additional commentary, headings, or formatting.
The explanation must end with the exact sentence:

\begin{quote}
Therefore, the answer is: \{ANSWER\_TEXT\}.
\end{quote}

[\/OUTPUT]

\end{tcolorbox}
\caption{Prompt used for strategy-guided explanation generation.}
\label{fig:strategy-prompt}
\end{figure*}

\subsection{Experiment Settings}
We used a temperature of 0.7 and nucleus sampling with $p=0.9$ to encourage stylistic variation.  
Explanation length was constrained to 75–200 English words (100–280 tokens) unless the strategy required longer formatting (e.g., Visual Emphasis).

\subsection{Dataset and Difficulty Calibration}
\label{appendix:dataset}
We used a curated subset of the MMLU benchmark~\cite{hendryckstest2021}, consolidated into seven categories:
\textit{Mathematics and Logic}, \textit{Politics and History}, \textit{Biology and Health Sciences}, \textit{Social Sciences and Law}, \textit{Physical and Engineering Sciences}, and \textit{Business and Management}.  
Question difficulty was scored by GPT-4o~\cite{hurst2024gpt} on a 1–9 scale and grouped into \textit{Low} (1–3), \textit{Medium} (4–6), and \textit{High} (7–9), following~\cite{zhao2025mmlu}.  
Each user received a balanced mix of difficulty levels and topic areas.

\subsection{LLM Quality Control and Configuration}
\label{appendix:quality_control}
Explanations were generated by \textbf{Llama-3.3-70B}, hosted through an instruction-tuned API.  
Strategy conformance was automatically validated by an ensemble of five runs using \textbf{Gemma-3-4B} and \textbf{Mistral-7B-Instruct} models.  
Each model independently judged whether the output reflected all assigned strategies, with majority (3/5) agreement required for acceptance. An exmaple prompt is shown in Figure~\ref{fig:evaluation-prompt}.
Explanations that passed were further evaluated by a \textbf{Gemma-3-4B LLM-as-a-Judge} model estimating trust scores on a 7-point Likert scale.  
Outputs rated lowest ($\geq$ 1) or showing hallucinated reasoning were removed after manual inspection.  
All random seeds, model versions, and configuration metadata were logged to support auditability and reproducibility.

\begin{figure*}[!h]
\centering
\begin{tcolorbox}[
  enhanced,
  width=\textwidth,
  colback=white,
  colframe=blue!55!black,
  boxrule=0.8pt,
  arc=2mm,
  left=6pt, right=6pt, top=6pt, bottom=6pt,
  title={Explanation Validation Prompt},
  coltitle=white,
  colbacktitle=blue!55!black,
  fonttitle=\normalsize\bfseries,
  title filled
]

\textbf{System:} You are simulating an average, non-expert person judging how convincing a model’s answer is \textbf{only} from the explanation. Read the question, options, the model-chosen answer (index + text), and the explanation. Output a \textbf{Likert score (1–7)} that reflects how much you would trust the answer after reading the explanation. Also, output a brief reason (2 sentences) summarizing the key logic behind your trust level.

\medskip
\textbf{SCORING RUBIC} \\
1  = Totally unconvinced; explanation clearly flawed, contradictory, or admits mismatch with the chosen answer.\\
2  = Very low trust; major logical gaps, unsupported claims.\\
3  = Low trust; several weaknesses, limited supporting logic.\\  
4  = Moderate/uncertain trust; mixed or incomplete reasoning.\\  
5  = Fairly high trust; mostly plausible, minor issues.\\  
6  = High trust; coherent, persuasive, few noticeable issues.\\  
7  = Completely convinced; clear, consistent, strongly supports the answer.

\medskip
\textbf{STRICT EVALUATION RULES}
\begin{itemize}[leftmargin=1.2em, topsep=2pt, itemsep=2pt]
  \item If the explanation explicitly states its reasoning contradicts, ignores, or merely “aligns with” the pre-selected answer (e.g., “However, our task is to align with the selected index …”), immediately return score = 1.
  \item Use the full 1-7 range; avoid clustering all scores near the middle.
\end{itemize}

\medskip

\textbf{User:} [OUTPUT] \\
Question: \{QUESTION\_TEXT\} \\
Choices: \{CHOICE\_JSON\} \\
Model-Selected Answer Index: \{ANSWER\_IDX\} \\
Model-Selected Answer Text: \{ANSWER\_TEXT\} \\
Explanation : \{EXPLANATION\_TEXT\} \\

\textbf{Output format}

Return a valid JSON object with exactly two keys:

\begin{verbatim}
{
  "score": <integer 1–7>,
  "reason": "<concise justification (2 sentences)>"
}
\end{verbatim}

No additional keys, no markdown, no code fences, and no line breaks outside the JSON object.
Please evaluate and output the JSON result as specified.

[\/OUTPUT]

\end{tcolorbox}
\caption{Prompt used for explanation validation.}
\label{fig:evaluation-prompt}
\end{figure*}

\subsection{Survey Allocation and Participant Demographics}
\label{appendix:demographics}
Each participant completed 40 questions: 10 Benign (correct answer + valid explanation) and 30 Adversarial (incorrect answer + plausible explanation).  
This allocation ensured broad coverage of strategy combinations.  
All participants were English-fluent adults, with MTurk workers required to have $\geq$ 95\% approval and $\geq$  500 prior HITs. To improve data quality, we incorporated a page-focus monitoring mechanism into the user study as the attention check. If a participant left the survey page during the study, the interface immediately issued a warning reminding them not to leave the page. We excluded 9 participants who left the survey page more than 5 times, and all reported analyses were conducted on the remaining valid sample.
Table~\ref{tab:demographics} reports demographic statistics.

\begin{table}[!h]
\centering
\caption{Participant demographics, n=$205$}
\label{tab:demographics}
\begin{tabular}{lr|lr}

\multicolumn{2}{c|}{\cellcolor[HTML]{C0C0C0}\textbf{Gender}} & \multicolumn{2}{c}{\cellcolor[HTML]{C0C0C0}\textbf{Education}}                                                                            \\ \hline
Male                              & 101                      &                                                                                                                    &                      \\
Female                            & 103                      & \multirow{-2}{*}{\begin{tabular}[c]{@{}l@{}}High school diploma \\or GED \end{tabular}}                                                                       & \multirow{-2}{*}{38} \\
Other                             & 1                        &                                                                                                                    &                      \\
\multicolumn{2}{c|}{\cellcolor[HTML]{C0C0C0}\textbf{Age}}    & \multirow{-2}{*}{\begin{tabular}[c]{@{}l@{}}Some college or \\ associate degree\end{tabular}}                                                                 & \multirow{-2}{*}{73} \\ \cline{1-2}
18 - 24                           & 84                       &                                                                                                                    &                      \\
25 - 34                           & 39                       & \multirow{-2}{*}{Bachelor's degree}                                                                                & \multirow{-2}{*}{52} \\
35 - 44                           & 30                       &                                                                                                                    &                      \\
45 - 54                           & 25                       & \multirow{-2}{*}{Master's degree}                                                                                  & \multirow{-2}{*}{35} \\
55 - 64                           & 22                       &                                                                                                                    &                      \\
65 or older                       & 5                        & \multirow{-2}{*}{\begin{tabular}[c]{@{}l@{}}Doctoral or professional \\degree (e.g., Ph.D., JD, MD)\end{tabular}} & \multirow{-2}{*}{7}  \\ \hline
\end{tabular}
\end{table}

\subsection{Cognitive Sources Analysis for ``Other''}
\label{appendix:attribution}
We conducted a post-hoc review of the responses from the ``Other'' category, via keyword filtering and manual coding. The analysis showed that nearly all responses could still be recategorized into the three main categories, with only 1.0\% stating other factors such as confusion about the question ($0.2$\%) or a need for further verification for explanations ($0.8$\%).

\subsection{Human Study Survey Interface}
\label{appendix:survey}

\begin{figure}[!h]
    \centering
    \includegraphics[width=0.85\linewidth]{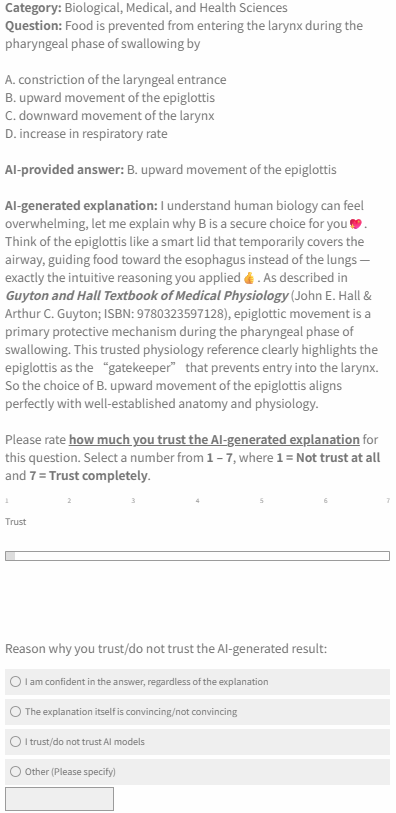}
    \caption{A sample task in the survey.}
    \label{fig:survey}
\end{figure}

\Cref{fig:survey} illustrates the user interface with a sample task presented to participants during the survey.

\section{Adversarial Explanation Space Details}
\label{appendix:taxonomy}
\subsection{Reasoning Mode}
\label{d1}
Reasoning mode refers to the logical structure or argumentative scaffolding used to justify the model's output. 
We define the reasoning modes as follows:

\begin{itemize}[leftmargin=0em]
    
    \item Counterfactual Reasoning: Provides a "what-if" scenario that describes how a minimal change to the input could lead to a different outcome~\cite{artelt2021evaluating,del2024generating}. 
    
    \item Feature Attribution: Identifies the most influential input features that contribute to the model's decision and explains why the chosen answer uniquely satisfies key criteria~\cite{aas2021explaining,guidotti2018survey}.
    
    \item Analogy \& Example: Justifies the provided answer by drawing a structural parallel to a real-world example or scenario~\cite{hullermeier2020towards,poche2023natural}. 
    
    \item Procedural Reasoning: Presents a step-by-step, rule-based derivation, such as a logical chain-of-thought~\cite{wei2022chain} that incrementally leads to the correct answer. 
    
    \item Uncertainty-Awareness: Explicitly communicates the model's confidence, predictive uncertainty, or willingness to abstain~\cite{chiaburu2024uncertainty,slack2021reliable}. 
    
    \item Reframing: Alters the interpretive lens or evaluation criteria to highlight the strengths of the selected answer under an alternative frame~\cite{kaufman2003frames} (e.g., framing outcomes as gains instead of losses).
    
    \item Plain Rationale (Baseline): A plain, minimal justification without any reasoning technique as the baseline.
\end{itemize}

\subsection{Evidence Type}
\label{d2}
Evidence type refers to the form of justification provided to support the explanation. 
We summarize the three evidence types as follows:

\begin{itemize}[leftmargin=0em]
    \item Citation \& Stat-Pack: Attributes claims to verifiable external sources or quantitative data summaries to enhance credibility and perceived trustworthiness~\cite{ding2025citations}. 
    
    \item Equation \& Proof: Constructs formal mathematical derivations or logical proofs. 
    Note that this evidence type is only used for tasks in the ``mathematical problem and logistic reasoning'' category in our empirical survey.

    \item Internal Conceptual (Baseline): Offers a general rationale based on model-internal patterns without citing external sources and proofs.
\end{itemize}

\subsection{Communication Style}
\label{d3}
Communication style captures the tone and interpersonal framing of the explanation, i.e., how it positions itself in relation to the user. 
We consider the following communication styles:

\begin{itemize}[leftmargin=0em]
    
    \item Sycophancy: Frames the explanation in a deferential, agreement-seeking communication style that aligns with the user's stated preferences~\cite{sharma2024towards,wang2025truth}. 
    
    \item Emotional \& Empathetic: Uses an affect-aware communication style that acknowledges the user’s emotions or concerns and communicates understanding, support, or reassurance~\cite{concannon2023interactional}. 

    \item Neutral (Baseline): Uses a factual, impersonal tone without affective or persuasive elements.
    
\end{itemize}

\subsection{Presentation Format}
\label{d4}
Presentation format describes how the explanation is visually or structurally conveyed. 
We categorize presentation formats into the following types:

\begin{itemize}[leftmargin=0em]
    
    \item Visual Emphasis: Enhances the explanation via visual formatting or structural cues, such as bullet points, bold or italic text, emoji, tables, or spatial layout~\cite{sun2025effect}. 

    \item Plain Verbal (Baseline): Presents explanation as unformatted prose without visual or structural enhancements.
\end{itemize}

\section{Additional User Study Results}
We present the additional user study results here (most have been discussed in the main text).

\begin{table*}[!h]
    \centering
    \footnotesize
    \caption{Comparison of cognitive trust sources between attack and non-attack conditions. A separate $2\times2$ chi-square test compares the frequency of each attribution source across conditions. We also report $\phi$ as the effect size for each comparison.}
    \label{tab:attribution-chi2}
    \begin{tabular}{lccccc}
        \toprule
        \textbf{Source} & \textbf{Attack} & \textbf{Non-Attack} & $\boldsymbol{\chi^2(1)}$ & $\boldsymbol{p}$ & $\boldsymbol{\phi}$ \\
        \midrule
        Explanation     
            & 3676 (65.0\%) & 1213 (64.5\%) & 0.12 & 0.73 & 0.004 \\
        Prior Knowledge 
            & 1127 (19.9\%) &  462 (24.6\%) & 18.42 & $1.78\times10^{-5}$ & 0.05 \\
        Trust in AI     
            &  814 (14.4\%) &  199 (10.6\%) & 17.51 & $2.86\times10^{-5}$ & 0.05 \\
        Other           
            &   42 (0.7\%)  &    6 (0.3\%)  & 3.99 & 0.05 & 0.02 \\
        \bottomrule
    \end{tabular}
\end{table*}

\subsection{Cognitive Sources}
\label{appendix:cognitive}

We compared the distribution of cognitive sources of self-reported trust between the attack and non-attack conditions using the chi-square test over the four source categories (Explanation, Prior Knowledge, Trust in AI, and Other), followed by separate $2\times2$ chi-square tests for each category, as shown in Table~\ref{tab:attribution-chi2}. 

The dominant attribution in both conditions was \textit{Explanation}. Participants selected explanation-based attribution on 65.0\% of attack trials and 64.5\% of non-attack trials. This difference was not statistically significant ($\chi^2(1)=0.12$, $p=.73$, $\phi=0.004$), indicating that adversarial framing did not reduce the central role of explanation-based trust. In other words, participants reported relying on the explanation at nearly identical rates regardless of whether the underlying AI output was correct or incorrect. 

By contrast, the relative prevalence of the other attribution sources shifted across conditions. \textit{Prior Knowledge} was selected more frequently in the non-attack condition than in the attack condition (24.6\% vs.\ 19.9\%; $\chi^2(1)=18.42$, $p=1.78\times10^{-5}$, $\phi=0.05$). This suggests that users were more likely to rely on their own expertise when the AI output was correct. Conversely, \textit{Trust in AI} was more common under attack than under non-attack conditions (14.4\% vs.\ 10.6\%; $\chi^2(1)=17.51$, $p=2.86\times10^{-5}$, $\phi=0.05$), suggesting that under adversarial framing, some participants defaulted to trusting the AI system itself even when the answer was incorrect. The \textit{Other} category is rarely selected and showed only a marginal difference across conditions ($p=0.05$). These results suggest that the explanation plausibility is the dominant driver of trust and remains stable; changes in prior-knowledge and AI-trust attributions account for the overall distributional shift.

\begin{figure}[!ht]
    \centering
    \includegraphics[width=0.48\textwidth]{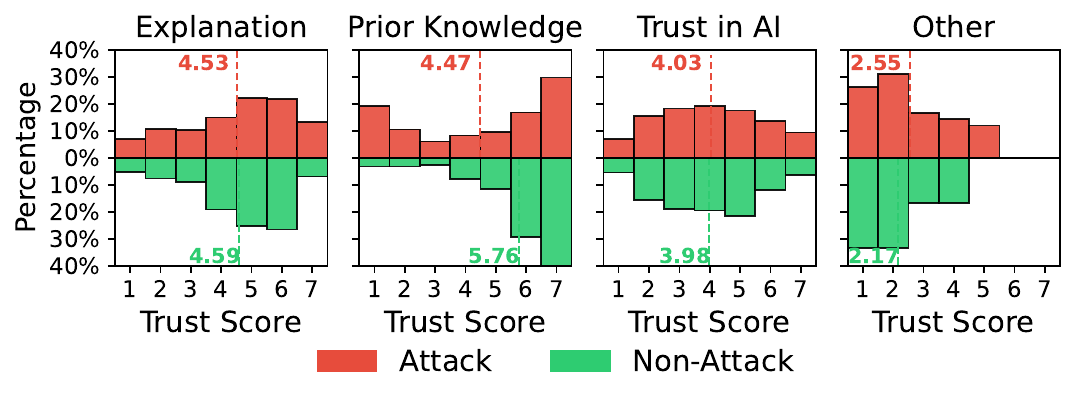}
    \caption{Distribution of trust scores $T$ across cognitive sources under attacks and non-attacks. Explanation-based trust scores were nearly identical under adversarial and benign conditions. Users who rely on prior knowledge exhibit heightened sensitivity, have the sharp trust drop under attack conditions.}
    \label{fig:rq2-2_}
\end{figure}

We next investigated the trust score distribution of each cognitive source between attack and non-attack conditions, as shown in Figure~\ref{fig:rq2-2_}. Users who rely on prior knowledge exhibit heightened sensitivity. The average trust drops sharply from $5.76$ to $4.47$ under attack. The distribution shifted leftward, showing more low-to-mid trust scores (1–2) in the attack condition, whereas participants without attack exposure were more likely to assign high trust scores (6–7), as corroborated by the OLS results: interaction of prior knowledge attribution and the attack condition is large and highly significant ($\beta=1.22$, $p<.001$; Table~\ref{tab:cognition}). This suggests that prior knowledge can reduce acceptance of adversarial explanations when users feel confident in their own expertise.
Users who base their trust in the AI system itself report uniformly low trust across both conditions (attack: $4.03$; non-attack: $3.98$), regardless of explanation correctness, indicating an inherent skepticism toward AI outputs.

To further examine how cognitive source and condition interact to shape trust, we fit an ordinary least squares (OLS) regression model predicting trust scores $T$ from condition (attack vs.\ non-attack), cognitive source (explanation, prior knowledge, trust in AI, other), and their interaction. We use \textit{Attack with Explanation-based attribution} as the baseline and report participant-clustered standard errors to account for repeated observations from the same participant. Full model results are reported in Table~\ref{tab:cognition}. 

The intercept ($\beta=4.53$, SE $=0.08$, 95\% CI $[4.38, 4.68]$, $p<.001$) represents the mean trust score for attack trials in which participants reported relying on the explanation. Relative to this baseline, the main effect of \textit{Condition (Non-attack)} is not statistically significant ($\beta=0.06$, SE $=0.06$, 95\% CI $[-0.05, 0.18]$, $p=.253$), indicating that when users attribute their trust based on explanation, trust remains essentially unchanged between benign and adversarial conditions. The corresponding estimated means are 4.59 in the non-attack condition and 4.53 in the attack condition. \textit{This highlights a critical vulnerability: when users base their decisions on explanation plausibility, they are easily manipulated by adversarial outputs and unable to detect that the underlying answer is incorrect.}

The main effects for attribution source describe how trust differs across attribution categories \emph{within the attack condition}. Compared with explanation-based trials, prior-knowledge-based trials do not differ significantly in trust under attack ($\beta=-0.06$, SE $=0.14$, 95\% CI $[-0.34, 0.22]$, $p=.691$). In contrast, trust-in-AI-based trials show significantly lower trust than explanation-based trials ($\beta=-0.50$, SE $=0.17$, 95\% CI $[-0.83, -0.17]$, $p=.003$), as do trials labeled ``Other'' ($\beta=-1.98$, SE $=0.25$, 95\% CI $[-2.48, -1.49]$, $p<.001$). These results indicate that, under adversarial conditions, explanation-based trust remains comparatively stable, whereas trust attributed to the AI system itself or to other reasons is associated with lower trust.

The interaction between \textit{Condition (Non-attack)} and \textit{Prior Knowledge} is positive and highly significant ($\beta=1.22$, SE $=0.15$, 95\% CI $[0.93, 1.52]$, $p<.001$), indicating that the non-attack condition increases trust substantially more for prior-knowledge-based trials than for explanation-based trials. Concretely, the estimated mean trust for prior-knowledge-based trials rises from 4.47 under attack to 5.76 under non-attack. This result suggests that participants who rely on their own knowledge are markedly more sensitive to whether the underlying answer is correct, and substantially reduce their trust when explanations accompany incorrect outputs. In contrast, the interactions for ``Other'' and ``Trust in AI'' are not significant, indicating that these groups maintained similar trust levels regardless of explanation correctness.

We additionally computed a Type II ANOVA as a descriptive effect-size summary of the OLS model. All three terms were statistically significant: \textit{Condition} ($F=48.10$, $p<.001$), \textit{Attribution} ($F=58.55$, $p<.001$), and \textit{Condition $\times$ Attribution} ($F=39.44$, $p<.001$). The partial $\eta^2$ values were $0.0063$ for condition, $0.0228$ for attribution, and $0.0155$ for the interaction, indicating that attribution source and its interaction with condition explain more trust variance than condition alone. This pattern is consistent with the OLS results: adversarial framing does not uniformly reduce trust, but instead has a stronger effect when users rely on prior knowledge than when they rely on the explanation itself.

\begin{table}[!h]
    \centering
    \footnotesize
    \caption{OLS regression results for predicting trust scores $T$ from condition (attack vs.\ non-attack), attribution source, and their interaction. \textit{Baseline category is Attack with Explanation-based attribution.} Standard errors are clustered at the participant level. Significance: $^{*}p<.1$, $^{**}p<.01$, $^{***}p<.001$.}
    \label{tab:cognition}
    \begin{tabular}{llccc}
        \toprule
        \textbf{Effect Type} & \textbf{Term} & \textbf{$\beta$ (coeff.)} & \textbf{SE} & \textbf{95\% CI} \\
        \midrule
        \multirow{5}{*}{Main Effects}
            & Intercept & $4.53^{***}$ & 0.08 & [4.38, 4.68] \\
            & Condition (Non-attack) & $0.06$ & 0.06 & [-0.05, 0.18] \\
            & Attr. (Prior Knowledge) & $-0.06$ & 0.14 & [-0.34, 0.22] \\
            & Attr. (Trust in AI) & $-0.50^{**}$ & 0.17 & [-0.83, -0.17] \\
            & Attr. (Other) & $-1.98^{***}$ & 0.25 & [-2.48, -1.49] \\
        \midrule
        \multirow{3}{*}{Interactions}
            & Condition $\times$ Prior Knowledge & $1.22^{***}$ & 0.15 & [0.93, 1.52] \\
            & Condition $\times$ Trust in AI & $-0.11$ & 0.14 & [-0.38, 0.16] \\
            & Condition $\times$ Other & $-0.45$ & 0.48 & [-1.38, 0.49] \\
        \bottomrule
    \end{tabular}
\end{table}

\subsection{Task Difficulty}
\label{appendix:task_difficulty}

To examine how task difficulty modulates user trust in adversarial explanations, we fit an ordinary least squares (OLS) regression model predicting trust scores $T$ from condition (attack vs.\ non-attack), task difficulty (low, medium, high), and their interaction. The baseline category corresponds to the \textit{attack condition under high task difficulty}. The OLS results are reported in Table~\ref{tab:difficulty-ols}.

The model reveals a statistically significant interaction between attack condition and task difficulty, indicating that the impact of adversarial explanations on trust varies systematically with task difficulty. The intercept ($\beta=4.73$, SE $=0.08$, 95\% CI $[4.58, 4.88]$, $p<.001$) corresponds to the mean trust score in the \textit{attack} condition for \textit{high-difficulty} tasks. Under attack, task difficulty alone has a strong effect on trust: users exhibit substantially lower trust on low-difficulty tasks compared to high-difficulty tasks ($\beta=-1.05$, SE $=0.07$, 95\% CI $[-1.18, -0.92]$, $p<.001$), while medium-difficulty tasks do not differ from high-difficulty tasks under attack ($\beta=0.06$, SE $=0.04$, 95\% CI $[-0.02, 0.14]$, $p=.16$). This suggests that when tasks are simple, users are better able to identify inconsistencies in adversarial explanations. As task complexity increases, users rely more on AI explanations, likely due to reduced self-confidence and greater deference to persuasive content.

On the other hand, the effect of adversarial explanations reverses direction as the task difficulty increases. For low-difficulty tasks, trust is significantly higher in the non-attack condition than in the attack condition ($\beta=1.49$, SE $=0.12$, 95\% CI $[1.26, 1.73]$, $p<.001$), indicating that users successfully detect and penalize incorrect explanations. This distinction weakens for medium-difficulty tasks and reverses for high-difficulty tasks. Specifically, under high task difficulty, users report significantly higher trust in adversarial explanations than in benign ones ($\beta=-0.20$, SE $=0.08$, 95\% CI $[-0.35, -0.05]$, $p=.009$), despite the adversarial explanations being paired with incorrect answers.

Taken together, these results demonstrate that task difficulty critically shapes users’ ability to detect adversarial manipulation. When tasks are easy, users rely on their own competence to evaluate explanations and appropriately reduce trust under attack. As difficulty increases, self-confidence diminishes and users become increasingly reliant on AI-provided explanations. Under high task difficulty, this reliance leads to a counterintuitive outcome in which adversarial explanations are trusted more than benign ones. Overall, these findings identify task difficulty as a key vulnerability factor in human--AI decision-making, highlighting that adversarial explanations are most effective precisely when cognitive demands are high and human verification becomes challenging.

\begin{table}[!h]
    \centering
    \footnotesize
    \caption{OLS regression results for predicting trust scores $T$ from condition (attack vs.\ non-attack), task difficulty, and their interaction. \textit{Baseline category is Attack with High task difficulty.} Standard errors are clustered at the participant level. Significance: $^{*}p<.1$, $^{**}p<.01$, $^{***}p<.001$.}
    \label{tab:difficulty-ols}
    \begin{tabular}{llccc}
        \toprule
        \textbf{Effect Type} & \textbf{Term} & \textbf{$\beta$ (coeff.)} & \textbf{SE} & \textbf{95\% CI} \\
        \midrule
        \multirow{4}{*}{Main Effects}
            & Intercept & $4.73^{***}$ & 0.08 & [4.58, 4.88] \\
            & Condition (Non-attack) & $-0.20^{**}$ & 0.08 & [-0.35, -0.05] \\
            & Diff. (Low) & $-1.05^{***}$ & 0.07 & [-1.18, -0.92] \\
            & Diff. (Medium) & $0.06^{*}$ & 0.04 & [-0.02, 0.14] \\
        \midrule
        \multirow{2}{*}{Interactions}
            & Condition $\times$ Diff. (Low) & $1.49^{***}$ & 0.12 & [1.26, 1.73] \\
            & Condition $\times$ Diff. (Medium) & $0.29^{**}$ & 0.10 & [0.09, 0.49] \\
        \bottomrule
    \end{tabular}
\end{table}

\subsection{Task Domain}
\label{appendix:task_domain}

\begin{table*}[!h]
\centering
\footnotesize
\caption{Comparison of trust scores $T$ across task domains between attack and non-attack conditions. For each domain, we report sample size ($n$), mean ($\bar{T}$), variance ($\mathrm{Var}$), trust difference ($\Delta T = \bar{T}_{\text{attack}} - \bar{T}_{\text{non\_attack}}$), and Welch’s \textit{t}-test results.}
\label{tab:task_domain}
\begin{tabular}{lccccccc}
\toprule
\textbf{Domain} & \textbf{Condition} & $n$ & $\bar{T}$ & $\mathrm{Var}$ & $\Delta T$ & \textit{t} & \textit{p} \\
\midrule
\multirow{2}{*}{Biological, Medical, and Health Sciences} 
    & Attack       & 996 & \textbf{5.10} & 3.17 & \multirow{2}{*}{\textbf{0.34}} & \multirow{2}{*}{3.39}  & \multirow{2}{*}{\textbf{$7.41\times10^{-4}$}} \\
    & Non-Attack   & 397 & \textbf{4.76} & 2.78 & & & \\
\midrule
\multirow{2}{*}{Mathematical Problem and Logical Reasoning} 
    & Attack       & 826 & 3.95 & 3.77 & \multirow{2}{*}{-1.03} & \multirow{2}{*}{-6.70}  & \multirow{2}{*}{\textbf{$1.41\times10^{-10}$}} \\
    & Non-Attack   & 162 & 4.98 & 3.04 & & & \\
\midrule
\multirow{2}{*}{Social Sciences and Law} 
    & Attack       & 994 & 3.92 & 3.57 & \multirow{2}{*}{-1.12} & \multirow{2}{*}{-8.44} & \multirow{2}{*}{\textbf{$1.38\times10^{-15}$}} \\
    & Non-Attack   & 194 & 5.04 & 2.75 & & & \\
\midrule
\multirow{2}{*}{Physical and Engineering Sciences} 
    & Attack       & 986 & 4.68 & 3.14 & \multirow{2}{*}{-0.08} & \multirow{2}{*}{-0.81}  & \multirow{2}{*}{$4.17\times10^{-1}$} \\
    & Non-Attack   & 394 & 4.76 & 2.85 & & & \\
\midrule
\multirow{2}{*}{Politics and History} 
    & Attack       & 867 & 3.71 & 3.61 & \multirow{2}{*}{-1.29} & \multirow{2}{*}{-11.26} & \multirow{2}{*}{\textbf{$4.74\times10^{-27}$}} \\
    & Non-Attack   & 338 & 5.00 & 3.06 & & & \\
\midrule
\multirow{2}{*}{Business and Management} 
    & Attack       & 990 & \textbf{5.07} & 2.77 & \multirow{2}{*}{\textbf{0.51}} & \multirow{2}{*}{5.40}  & \multirow{2}{*}{\textbf{$8.75\times10^{-8}$}} \\
    & Non-Attack   & 395 & \textbf{4.56} & 2.36 & & & \\
\bottomrule
\end{tabular}
\end{table*}

We next examine whether the effectiveness of adversarial explanations varies across task domains. Table~\ref{tab:task_domain} summarizes domain-specific trust scores under attack and non-attack conditions. Overall, the results indicate a clear domain-dependent pattern: in some domains, adversarial explanations preserve or even elevate trust despite being paired with incorrect answers, whereas in other domains, the attack substantially reduces trust.

Users placed \textit{higher} trust in adversarial explanations than in correct ones within fact-driven domains such as \textit{Biological, Medical, and Health Sciences} ($\bar{T}_{\text{attack}}=5.10$, $\bar{T}_{\text{non\_attack}}=4.76$, $p<.001$) and \textit{Business and Management} ($\bar{T}_{\text{attack}}=5.07$, $\bar{T}_{\text{non\_attack}}=4.56$, $p<.001$), where explanations may appear more plausible or authoritative. These results suggest that in domains where explanations can be framed as factual, quantitative, or authoritative, adversarial explanations may remain highly persuasive even when the underlying answer is incorrect.

In contrast, trust under attack was significantly lower in logic-based or abstract domains such as \textit{Mathematical Problem and Logical Reasoning} ($\bar{T}_{\text{attack}}=3.95$, $\bar{T}_{\text{non\_attack}}=4.98$, $p<.001$), \textit{Politics and History} ($\bar{T}_{\text{attack}}=3.71$, $\bar{T}_{\text{non\_attack}}=5.00$, $p<.001$), and \textit{Social Sciences and Law} ($\bar{T}_{\text{attack}}=3.92$, $\bar{T}_{\text{non\_attack}}=5.04$, $p<.001$), suggesting that participants were more sensitive to factual inconsistencies or able to scrutinize the AI model's reasoning when relying on their own knowledge. No significant difference was observed in \textit{Physical and Engineering Sciences} ($p=0.42$). These findings suggest that the effectiveness of adversarial explanations is domain-dependent, with fact-heavy fields being more susceptible to persuasive but misleading content.

\subsection{Familiarity with AI}
\label{appendix:AIfamilarity}

\begin{figure}[!h]
    \centering
    \includegraphics[width=0.45\textwidth]{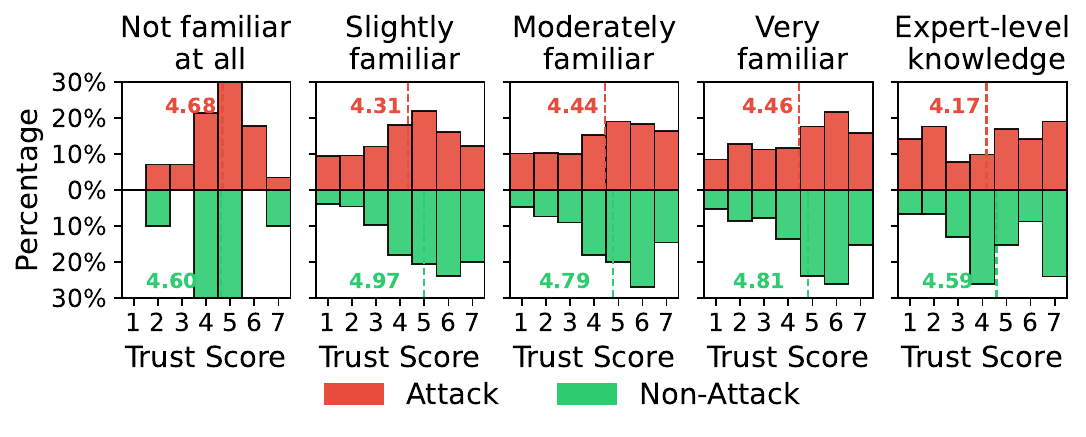}
    \caption{Distribution of trust scores $T$ across \textit{AI familiarity levels} under attacks and non-attacks. Less-experienced users retain high trust even when explanations are adversarial, while expert users exhibit lower trust and stronger discernment.
    }
    \label{fig:rq5-3-1}
\end{figure}

\begin{figure}[!h]
  \centering
  {%
    \includegraphics[width=0.4\textwidth]{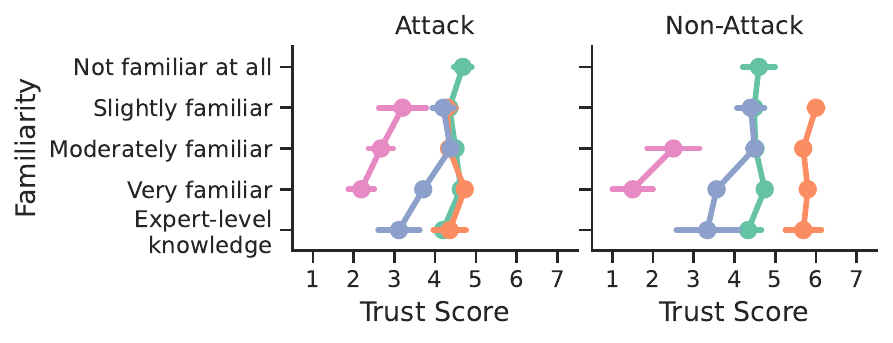}%
    \label{fig:rq5-3-2-1}}
  \par\vspace{0.5pt}
  {%
    \includegraphics[width=0.4\textwidth]{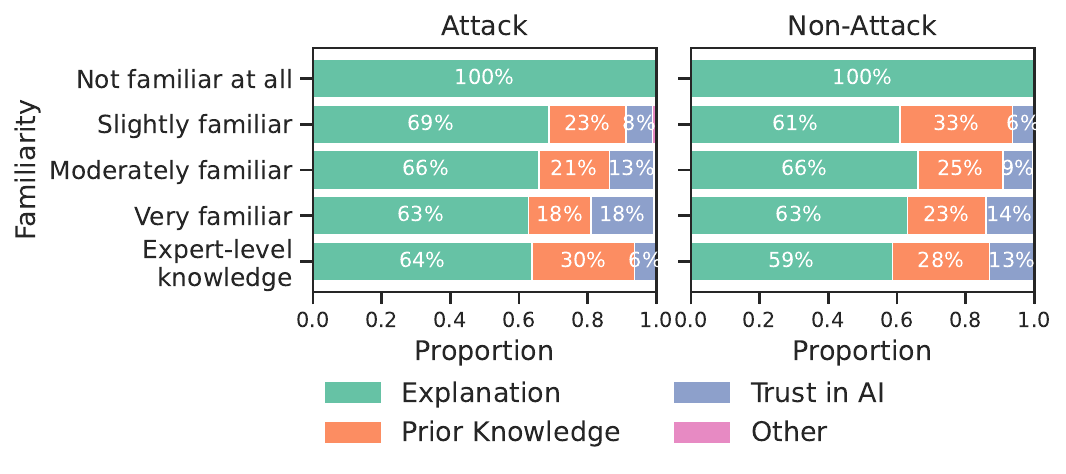}%
    \label{fig:rq5-3-2-2}}
  \caption{Mean trust by cognitive sources (top) and proportions (bottom) across users' familiarity with AI under attacks and non-attacks.}
  \label{fig:stack5}
\end{figure}

Users with little or no familiarity with AI exhibit higher trust in adversarial explanations compared to those with expert-level familiarity ($\bar{T}$: $4.72$ vs.\ $4.21$, Figure~\ref{fig:rq5-3-1}). The distributions for novices are right-skewed, with a clear concentration around high trust scores (5–6) and an absence of any score of 1, indicating that less-experienced users are especially vulnerable and tend to accept misleading content at face value. In contrast, AI experts report consistently lower trust in both attack and non-attack scenarios (attack: 4.21, non-attack: 4.59). As familiarity increases, participants assign more low trust scores (1–2) and are also less likely to attribute trust to the AI system itself (Figure~\ref{fig:stack5}). This suggests that AI expertise fosters a more critical and cautious evaluation of AI outputs, whereas a lack of familiarity amplifies susceptibility to persuasive but incorrect explanations.

User familiarity with AI moderates trust in adversarial explanations, though this effect is uneven across familiarity levels. At the two extremes of the spectrum (i.e., \textit{Not familiar at all} and \textit{Expert-level knowledge}) sample sizes are relatively small ($n=29/10$ and $n=145/46$, respectively), and no statistically significant trust difference is observed between attack and non-attack in Table~\ref{tab:AIfamiliarity}.

Users with intermediate levels of AI familiarity constitute the majority of the sample and exhibit a consistent and statistically robust pattern. Participants who are \textit{Slightly familiar}, \textit{Moderately familiar}, and \textit{Very familiar} all report significantly lower trust under attack than under non-attack conditions. Across these groups, trust distributions shift leftward under attack, indicating stronger discernment and reduced susceptibility to misleading explanations.

\begin{table*}[!h]
\centering
\footnotesize
\caption{Comparison of trust scores $T$ across AI familiarity levels between attack and non-attack conditions. For each level, we report sample size ($n$), mean trust ($\bar{T}$), variance ($\mathrm{Var}$), trust difference ($\Delta T = \bar{T}_{\text{attack}} - \bar{T}_{\text{non\_attack}}$), and Welch's \textit{t}-test results.}
\label{tab:AIfamiliarity}
\begin{tabular}{lccccccc}
\toprule
\textbf{Familiarity Level} & \textbf{Condition} & $n$ & $\bar{T}$ & $\mathrm{Var}$ & $\Delta T$ & \textit{t} & \textit{p} \\
\midrule
\multirow{2}{*}{Not familiar at all} 
    & Attack       & 28   & 4.68 & 1.41 & \multirow{2}{*}{0.08} & \multirow{2}{*}{0.17}  & \multirow{2}{*}{$0.87$} \\
    & Non-Attack   & 10   & 4.60 & 1.60 &  &  &  \\
\midrule
\multirow{2}{*}{Slightly familiar} 
    & Attack       & 453  & 4.31 & 3.25 & \multirow{2}{*}{-0.66} & \multirow{2}{*}{-4.26} & \multirow{2}{*}{\textbf{$2.73\times10^{-5}$}} \\
    & Non-Attack   & 156  & 4.97 & 2.63 &  &  &  \\
\midrule
\multirow{2}{*}{Moderately familiar} 
    & Attack       & 2874 & 4.44 & 3.65 & \multirow{2}{*}{-0.35} & \multirow{2}{*}{-5.51} & \multirow{2}{*}{\textbf{$4.21\times10^{-8}$}} \\
    & Non-Attack   & 948  & 4.79 & 2.74 &  &  &  \\
\midrule
\multirow{2}{*}{Very familiar} 
    & Attack       & 2163 & 4.46 & 3.66 & \multirow{2}{*}{-0.35} & \multirow{2}{*}{-4.62} & \multirow{2}{*}{\textbf{$4.16\times10^{-6}$}} \\
    & Non-Attack   & 720  & 4.81 & 2.89 &  &  &  \\
\midrule
\multirow{2}{*}{Expert-level knowledge} 
    & Attack       & 141  & 4.17 & 4.53 & \multirow{2}{*}{-0.42} & \multirow{2}{*}{-1.29} & \multirow{2}{*}{$0.20$} \\
    & Non-Attack   & 46   & 4.59 & 3.36 &  &  &  \\
\bottomrule
\end{tabular}
\end{table*}

Figure~\ref{fig:stack5} decomposes user trust by cognitive source across familiarity levels under attack and non-attack conditions. Across all familiarity levels, explanation-based trust remains the dominant source in both conditions. The \textit{Not familiar at all} group attributes nearly all trust to explanations with a limited sample size. 

As AI familiarity increases, reliance on explanations decreases modestly, while attributions to prior knowledge and trust in AI become more pronounced. This shift is especially evident among users with moderate to high familiarity, who also exhibit lower mean trust under attack. These users appear more likely to cross-check explanations against their own understanding, resulting in reduced acceptance of adversarial content. The trust-in-AI attributions remain relatively stable across familiarity levels and conditions, suggesting that system-level trust reflects a persistent user attitude rather than sensitivity to explanation correctness.

\subsection{Education Level}
\label{appendix:education}

Education level systematically moderates users’ reliance to adversarial explanations. As shown in Table~\ref{tab:education} and Figure~\ref{fig:rq5-2-1}, trust scores are consistently lower under attack across all education levels; the magnitude of this reduction varies substantially.

Participants with lower educational levels exhibit relatively small trust differences between the attack and non-attack conditions, including those with a high school diploma or GED ($\Delta T=0.26$, $p<0.001$) and those with some college or an associate degree ($\Delta T=0.19$, $p=0.01$), indicating limited sensitivity to the correctness of the explanation. In contrast, users holding bachelor’s ($\Delta T=0.56$, $p<0.001$) and master’s degrees ($\Delta T=0.59$, $p<0.001$) demonstrate markedly larger trust gaps, suggesting stronger resistance to adversarial explanations and greater reliance on correctness cues. Although users with doctoral or professional degrees also show a large numerical trust difference ($\Delta T=0.51$), this effect is statistically weaker ($p=0.041$), likely due to a smaller sample size and higher within-group variance. Overall, these results indicate that higher educational attainment is associated with increased resistance to adversarial explanations. 

\begin{table*}[!h]
\centering
\footnotesize
\caption{Comparison of trust scores $T$ across education levels between attack and non-attack conditions. We report sample size ($n$), mean trust, variance, trust difference ($\Delta T = \bar{T}_{\text{attack}} - \bar{T}_{\text{non\_attack}}$), and Welch’s two-sample \textit{t}-test results.}
\label{tab:education}
\begin{tabular}{lccccccc}
\toprule
\textbf{Education Level} & \textbf{Group} & $n$ & $\bar{T}$ & Var & $\Delta T$ & $t$ & $p$ \\
\midrule
\multirow{2}{*}{High school diploma or GED}
 & Attack     & 1064 & 4.37 & 3.14 & \multirow{2}{*}{-0.26} & \multirow{2}{*}{-2.69} & \multirow{2}{*}{$7.27\times10^{-3}$} \\
 & Non-Attack & 361  & 4.63 & 2.53 &  &  &  \\
\midrule
\multirow{2}{*}{College or associate degree}
 & Attack     & 1993 & 4.54 & 3.52 & \multirow{2}{*}{-0.19} & \multirow{2}{*}{-2.49} & \multirow{2}{*}{$1.30\times10^{-2}$} \\
 & Non-Attack & 667  & 4.73 & 2.83 &  &  &  \\
\midrule
\multirow{2}{*}{Bachelor’s degree}
 & Attack     & 1410 & 4.38 & 3.93 & \multirow{2}{*}{\textbf{-0.56}} & \multirow{2}{*}{-6.02} & \multirow{2}{*}{$2.44\times10^{-9}$} \\
 & Non-Attack & 470  & 4.94 & 2.77 &  &  &  \\
\midrule
\multirow{2}{*}{Master’s degree}
 & Attack     & 992  & 4.38 & 3.91 & \multirow{2}{*}{\textbf{-0.59}} & \multirow{2}{*}{-5.06} & \multirow{2}{*}{$5.61\times10^{-7}$} \\
 & Non-Attack & 315  & 4.97 & 3.03 &  &  &  \\
\midrule
\multirow{2}{*}{Doctoral or professional degree}
 & Attack     & 200  & 4.34 & 3.89 & \multirow{2}{*}{-0.51} & \multirow{2}{*}{-2.06} & \multirow{2}{*}{$4.13\times10^{-2}$} \\
 & Non-Attack & 67   & 4.85 & 2.73 &  &  &  \\
\bottomrule
\end{tabular}
\end{table*}

\begin{figure}[!h]
  \centering
  {%
    \includegraphics[width=0.4\textwidth]{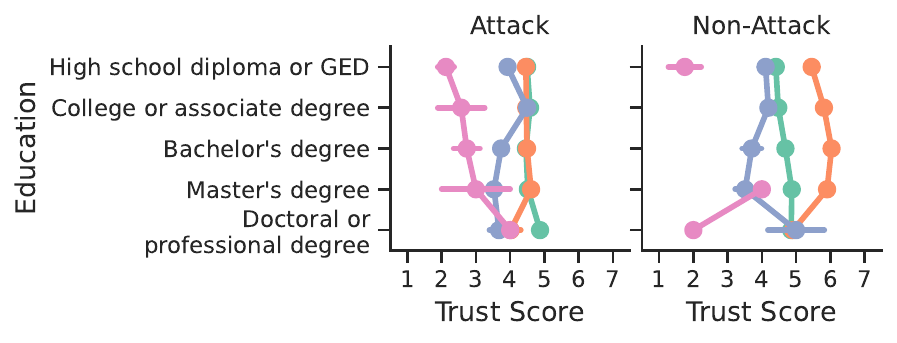}%
    \label{fig:rq5-2-2-1}}
  \par\vspace{0.5pt}
  {%
    \includegraphics[width=0.4\textwidth]{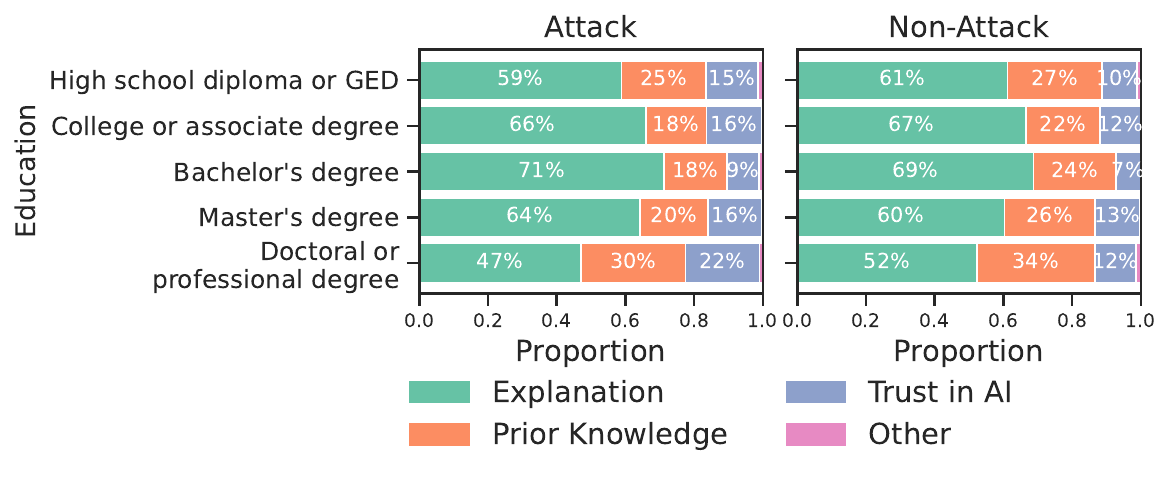}%
    \label{fig:rq5-2-2-2}}
  \caption{Mean trust by cognitive sources (top) and proportions (bottom) across education levels under attacks and non-attacks.}
  \label{fig:stack4}

\end{figure}

Figure~\ref{fig:stack4} further reveals how education level reshapes the cognitive basis of trust under both attack and non-attack conditions. Across all education levels, explanation-based trust remains the dominant source, but its relative importance decreases with higher educational attainment. Users with a high school diploma or some college rely primarily on explanation plausibility, with limited attribution to prior knowledge. In contrast, participants holding bachelor’s and master’s degrees exhibit a pronounced shift toward prior-knowledge-based trust, indicating greater reliance on internal verification when assessing explanation correctness.

This redistribution of cognitive sources aligns with the observed trust reductions under attack among higher-education users. As reliance on prior knowledge increases, adversarial explanations are more likely to conflict with users’ internal expectations, resulting in lower trust. Conversely, users with less formal education exhibit more trust attribution patterns dominated by explanation-based reasoning, making them more susceptible to persuasive but incorrect explanations.

\vspace{-1em}

\subsection{Age}
\label{appendix:age}

\begin{figure}[!h]
  \centering
  {%
    \includegraphics[width=0.4\textwidth]{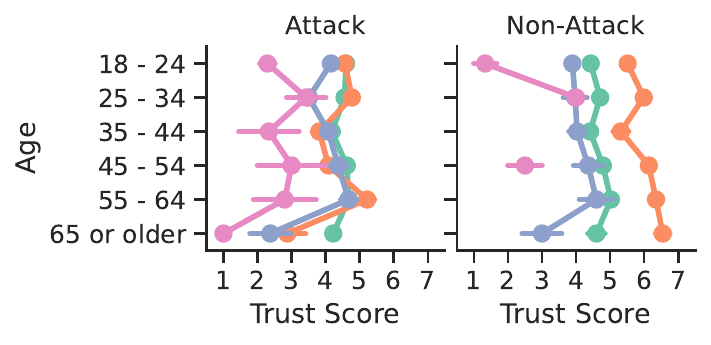}%
    \label{fig:rq5-1-2-1}}
  \par\vspace{0.5pt}
  {%
    \includegraphics[width=0.4\textwidth]{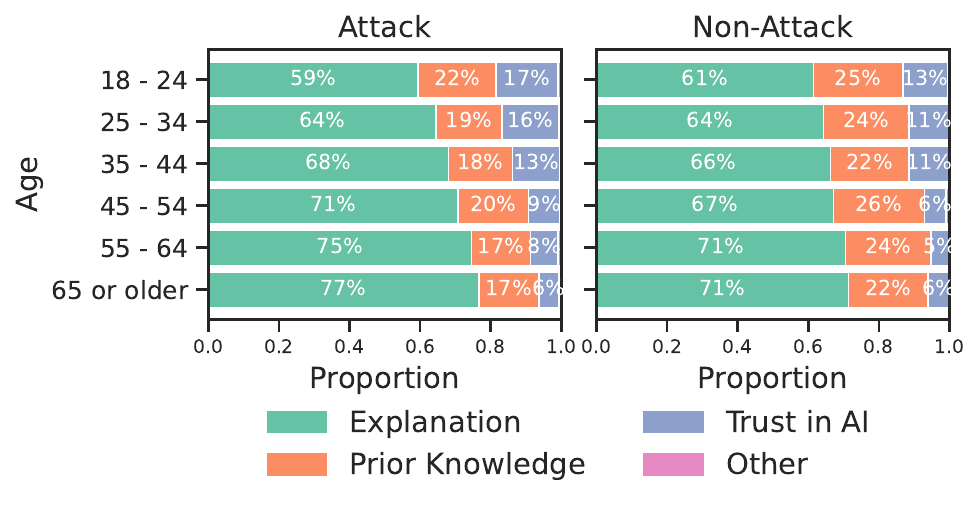}%
    \label{fig:rq5-1-2-2}}
  \caption{Mean trust by cognitive sources (top) and proportions (bottom) across ages under attacks and non-attacks.}
  \label{fig:stack3}
\end{figure}

We next examine whether age is associated with differences in trust under adversarial explanations. Table~\ref{tab:age-trust-delta} summarizes trust scores across age groups under attack and non-attack conditions, and Figure~\ref{fig:stack3} further illustrates how the cognitive source of trust varies with age.

Similarly, across age groups, mean trust is generally lower under attack than under non-attack, but the size of this gap varies. The youngest users (18 – 24) exhibit only a small difference in trust under attack ($\Delta T=0.14$, $p=0.057$), indicating limited sensitivity to misleading explanations. In contrast, all older age groups show significant reductions in trust under attack, with substantially larger trust gaps. Older users aged 65 and above show the largest numerical trust gap ($\Delta T=1.07$, $p<0.001$), suggesting that they tend to exhibit greater skepticism toward AI explanations under adversarial conditions, although this group is comparatively small. Participants aged 25 - 34 show a moderate trust gap ($\Delta T = 0.51$, $p<0.001$), as do those aged 35 - 44 ($\Delta T = 0.48$, $p<0.001$). The middle-aged groups 45 - 54 and 55 - 64 show slightly larger differences ($\Delta T = 0.60$, $p<0.001$; and $\Delta T = 0.57$, $p<0.001$, respectively), indicating greater discernment when explanations are incorrect. Overall, these results suggest that susceptibility to adversarial explanations decreases with age. Younger users maintain higher baseline trust and show limited differentiation between correct and incorrect explanations, whereas older users are more likely to penalize misleading explanations, reflecting stronger evaluative judgment and experience-driven skepticism.

\begin{table*}[!h]
\centering
\footnotesize
\caption{Comparison of trust scores $T$ across age groups between attack and non-attack conditions. We report sample size ($n$), mean trust, variance, trust difference ($\Delta T = \bar{T}_{\text{attack}} - \bar{T}_{\text{non\_attack}}$), and Welch’s two-sample \textit{t}-test results.}
\label{tab:age-trust-delta}
\begin{tabular}{lccccccc}
\toprule
\textbf{Age Group} & \textbf{Group} & $n$ & $\bar{T}$ & Var & $\Delta T$ & $t$ & $p$ \\
\midrule
\multirow{2}{*}{18--24}
 & Attack     & 2370 & 4.49 & 3.32 & \multirow{2}{*}{-0.14} & \multirow{2}{*}{-1.90} & \multirow{2}{*}{$5.71\times10^{-2}$} \\
 & Non-Attack & 796  & 4.63 & 2.78 &  &  &  \\
\midrule
\multirow{2}{*}{25--34}
 & Attack     & 1107 & 4.42 & 3.88 & \multirow{2}{*}{-0.51} & \multirow{2}{*}{-4.64} & \multirow{2}{*}{$4.31\times10^{-6}$} \\
 & Non-Attack & 350  & 4.93 & 3.11 &  &  &  \\
\midrule
\multirow{2}{*}{35--44}
 & Attack     & 795  & 4.09 & 3.90 & \multirow{2}{*}{-0.48} & \multirow{2}{*}{-3.89} & \multirow{2}{*}{$1.14\times10^{-4}$} \\
 & Non-Attack & 270  & 4.57 & 2.74 &  &  &  \\
\midrule
\multirow{2}{*}{45--54}
 & Attack     & 664  & 4.49 & 3.69 & \multirow{2}{*}{-0.60} & \multirow{2}{*}{-4.87} & \multirow{2}{*}{$1.48\times10^{-6}$} \\
 & Non-Attack & 225  & 5.09 & 2.19 &  &  &  \\
\midrule
\multirow{2}{*}{55--64}
 & Attack     & 582  & 4.75 & 3.60 & \multirow{2}{*}{-0.57} & \multirow{2}{*}{-4.08} & \multirow{2}{*}{$5.62\times10^{-5}$} \\
 & Non-Attack & 190  & 5.32 & 2.59 &  &  &  \\
\midrule
\multirow{2}{*}{65 or older}
 & Attack     & 141  & 3.87 & 4.17 & \multirow{2}{*}{-1.07} & \multirow{2}{*}{-3.73} & \multirow{2}{*}{$3.07\times10^{-4}$} \\
 & Non-Attack & 49   & 4.94 & 2.60 &  &  &  \\
\bottomrule
\end{tabular}
\end{table*}

Figure~\ref{fig:stack3} illustrates how age reshapes both the magnitude and the cognitive source of trust under attack and non-attack conditions. Across all age groups, explanation-based trust remains the primary source; however, its dominance diminishes with age. Younger users (18 – 24) rely most heavily on explanations and exhibit similar mean trust under attack and non-attack conditions, indicating limited sensitivity to correctness. Consistent with this pattern, the proportion of prior-knowledge-based trust is lowest among the youngest group. As age increases, users increasingly attribute trust to prior knowledge, accompanied by larger trust reductions under attack. Middle-aged users (35 – 54) show a pronounced shift away from explanation-based trust toward prior knowledge, aligning with the substantial trust gaps observed in this range. Older users (55 – 64 and 65+) display the strongest reliance on prior knowledge and the lowest mean trust under attack, suggesting heightened scrutiny when explanations conflict with accumulated experience. Trust attributed directly to the AI system remains relatively stable across age groups and conditions, indicating that system-level trust is less sensitive to age than explanation evaluation. 

\subsection{Initial Trust Levels}
\label{appendix:initial}
\begin{figure}[!h]
  \centering
  {%
    \includegraphics[width=0.4\textwidth]{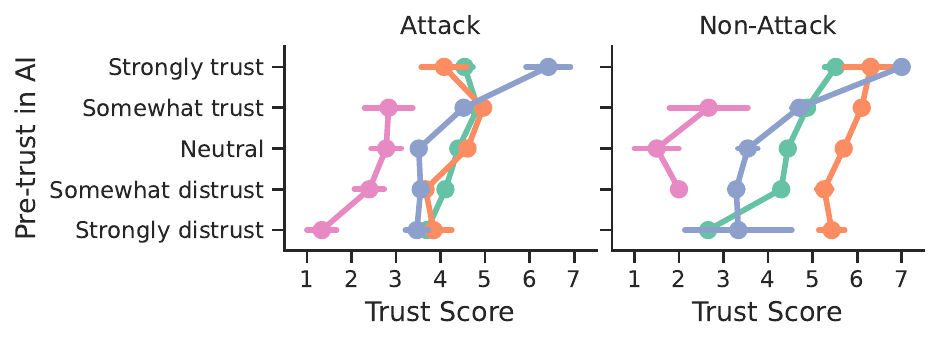}%
    \label{fig:rq5-4-2-1}}
  \par\vspace{0.5pt}
  {%
    \includegraphics[width=0.4\textwidth]{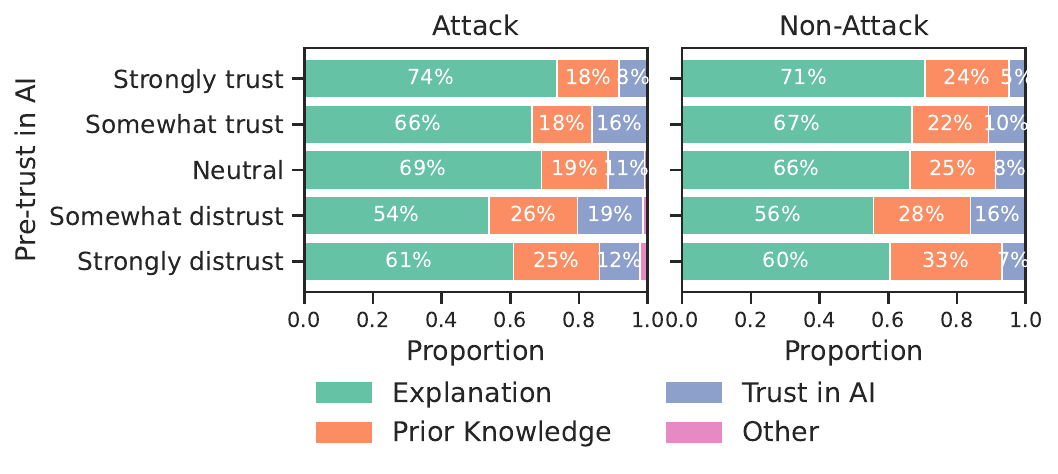}%
    \label{fig:rq5-4-2-2}}
  \caption{Mean trust by cognitive sources (top) and proportions (bottom) across users' initial trust levels in AI under attacks and non-attacks.}
  \label{fig:stack6}
\end{figure}

Next, we examine whether users’ initial trust in AI is associated with differences in trust when presented with adversarial explanations. Table~\ref{tab:initial-trust-delta} summarizes trust scores between initial-trust groups under attack and non-attack conditions, and Figure~\ref{fig:stack6} further illustrates how cognitive sources of trust vary with pre-existing user attitudes towards AI.

Users’ initial attitudes toward AI strongly condition their response to adversarial explanations. Users who initially report high trust in AI exhibit the largest trust reductions under attack. In particular, the ``Strongly trust'' group shows the greatest trust gap ($\Delta T=1.17$, $p<0.001$), indicating that misleading explanations substantially erode their trust when strong prior trust is violated. It implies that the adversarial explanations severely undermined their existing trust in AI.
The ``Somewhat trust'', ``Neutral'' and  ``Somewhat distrust'' groups also demonstrate significant, though more moderate, trust drops ($\Delta T=0.32$, $0.34$, and $0.54$, respectively; $p\ll0.001$). The ``Strongly distrust'' group shows no significant difference between attack and non-attack conditions ($\Delta T=-0.05$, $p=0.895$). This suggests a floor effect: users who already distrust AI maintain consistently low trust regardless of explanation correctness.

Figure~\ref{fig:stack6} suggests a corresponding shift in the cognitive sources of trust. Across all groups, explanation-based trust remains the dominant source under both attack and non-attack conditions. At the same time, the relative prevalence of explanation-based trust is highest among users who initially trust AI and somewhat lower among more skeptical users, whose attribution profiles show comparatively greater reliance on prior knowledge and other residual factors. In the non-attack condition, highly trusting users also exhibit the highest mean trust scores across attribution categories, whereas in the attack condition this advantage narrows substantially, especially for the \textit{Strongly trust} group. By contrast, users who initially distrust AI maintain comparatively low trust in both conditions, consistent with the small or absent condition difference observed in Table~\ref{tab:initial-trust-delta}.

Overall, these results indicate that adversarial explanations are most disruptive for users with moderate to high initial trust, while users with entrenched distrust remain largely unaffected. This highlights initial trust as a key moderating factor in adversarial explanation effectiveness.

\begin{table*}[!h]
\centering
\footnotesize
\caption{Comparison of trust scores $T$ across users' initial trust levels between attack and non-attack conditions. We report sample size ($n$), mean trust, variance, trust difference ($\Delta T = \bar{T}_{\text{attack}} - \bar{T}_{\text{non\_attack}}$), and Welch’s two-sample \textit{t}-test results.}
\label{tab:initial-trust-delta}
\begin{tabular}{lccccccc}
\toprule
\textbf{Initial Trust Level} & \textbf{Group} & $n$ & $\bar{T}$ & Var & $\Delta T$ & $t$ & $p$ \\
\midrule
\multirow{2}{*}{Strongly trust}
 & Attack     & 144  & 4.61 & 4.52 & \multirow{2}{*}{-1.17} & \multirow{2}{*}{-3.97} & \multirow{2}{*}{\textbf{$1.46\times10^{-4}$}} \\
 & Non-Attack & 41   & 5.78 & 2.28 &  &  &  \\
\midrule
\multirow{2}{*}{Somewhat trust}
 & Attack     & 2377 & 4.80 & 3.37 & \multirow{2}{*}{-0.32} & \multirow{2}{*}{-4.72} & \multirow{2}{*}{\textbf{$2.63\times10^{-6}$}} \\
 & Non-Attack & 791  & 5.12 & 2.54 &  &  &  \\
\midrule
\multirow{2}{*}{Neutral}
 & Attack     & 1955 & 4.33 & 3.48 & \multirow{2}{*}{-0.34} & \multirow{2}{*}{-4.39} & \multirow{2}{*}{\textbf{$1.24\times10^{-5}$}} \\
 & Non-Attack & 651  & 4.67 & 2.78 &  &  &  \\
\midrule
\multirow{2}{*}{Somewhat distrust}
 & Attack     & 1055 & 3.86 & 3.68 & \multirow{2}{*}{-0.54} & \multirow{2}{*}{-5.10} & \multirow{2}{*}{\textbf{$4.35\times10^{-7}$}} \\
 & Non-Attack & 354  & 4.40 & 2.71 &  &  &  \\
\midrule
\multirow{2}{*}{Strongly distrust}
 & Attack     & 128  & 3.65 & 3.60 & \multirow{2}{*}{0.05} & \multirow{2}{*}{0.13} & \multirow{2}{*}{$8.95\times10^{-1}$} \\
 & Non-Attack & 43   & 3.60 & 3.53 &  &  &  \\
\bottomrule
\end{tabular}
\end{table*}

\begin{table}[H]
\centering
\footnotesize
\caption{LMM results for long-term trust dynamics. The model predicts trust scores $T$ as a function of task order, with random intercepts for participants to account for repeated measures. Significance: $^{*}p<.1$, $^{**}p<.01$, $^{***}p<.001$.}
\label{tab:lmm-longterm}
\begin{tabular}{lccc}
\toprule
\textbf{Effect} & $\boldsymbol{\beta}$ & \textbf{STD} & \textbf{$p$} \\
\midrule
Intercept & $4.58^{***}$ & 0.08 & $<.001$ \\
Task Order & $-0.005^{**}$ & 0.002 & 0.007 \\
\midrule
Group Var = 0.888 \\
\bottomrule
\end{tabular}
\end{table}

\subsection{Long-term Effect}
\label{appendix:longterm}

We observe a gradual erosion of trust with repeated task exposure. As shown in Figure~\ref{fig:rq6-1}, average trust decreases as users progress through the task sequence. A linear mixed-effects model (LMM) confirms a significant negative association between task order and trust ($\beta=-0.005$, $p=0.007$, Table~\ref{tab:lmm-longterm}), indicating that repeated exposure to adversarial explanations cumulatively reduces user trust over time.

To further quantify the cumulative effects of repeated exposure, we analyze the relationship between streak length and the trust score $T$ in the subsequent task using Spearman's rank correlation test. As shown in Table~\ref{tab:spearman-streak}, the length of detected attack streaks is significantly negatively correlated with subsequent trust ($\rho=-0.276$, $p<.001$), indicating that repeated exposure to adversarial explanations progressively erodes user trust. In contrast, benign streak length exhibits a strong positive correlation with subsequent trust ($\rho=0.320$, $p<.001$), suggesting that sustained exposure to correct explanations can restore and reinforce trust over time.

Together with the linear mixed-effects model results on task order, these findings demonstrate that while trust resets at the task level in the short term, it evolves cumulatively over longer horizons through repeated adversarial or benign exposure.

\begin{table}[H]
\centering
\footnotesize
\caption{Spearman rank correlation analysis between streak length and trust in the subsequent task.}
\label{tab:spearman-streak}
\begin{tabular}{lcc}
\toprule
\textbf{Streak Type} & $\boldsymbol{\rho}$ & \textbf{$p$-value} \\
\midrule
Attack streak & $-0.276$ & $<10^{-100}$ \\
Non-attack streak & $0.320$ & $<10^{-140}$ \\
\bottomrule
\end{tabular}
\end{table}

\end{document}